\documentclass[letterpaper]{article} 
\usepackage{aaai2026} 
\usepackage{times} 
\usepackage{helvet} 
\usepackage{courier} 
\usepackage[hyphens]{url} 
\usepackage{graphicx} 
\usepackage{natbib} 
\usepackage{caption} 
\usepackage{algorithm}
\usepackage{algorithmic}
\usepackage{xcolor}
\usepackage{newfloat}
\usepackage{listings}
\DeclareCaptionStyle{ruled}{labelfont=normalfont,labelsep=colon,strut=off} 
\floatstyle{ruled}
\newfloat{listing}{tb}{lst}{}
\floatname{listing}{Listing}
\usepackage{subcaption}
\usepackage{booktabs} 
\usepackage{amsfonts,amsmath,amssymb,amsthm}
\newtheorem{theorem}{Theorem}
\newtheorem{lemma}[theorem]{Lemma}

\newtheorem{proposition}{Proposition}

\newcommand{\ours}{GCL-OT}

\title{GCL-OT: Graph Contrastive Learning with Optimal Transport for Heterophilic Text-Attributed Graphs}

\author {
 Yating Ren,
 Yikun Ban,
 Huobin Tan\thanks{Corresponding author}
}
\affiliations {
 Beihang University\\
 renyating@buaa.edu.cn, 
 yikunb@buaa.edu.cn, 
 thbin@buaa.edu.cn
}

\begin{document}

\maketitle

\begin{abstract}
Recently, structure–text contrastive learning has shown promising performance on text-attributed graphs by leveraging the complementary strengths of graph neural networks and language models. However, existing methods typically rely on homophily assumptions in similarity estimation and hard optimization objectives, which limit their applicability to heterophilic graphs. Although existing methods can mitigate heterophily through structural adjustments or neighbor aggregation, they usually treat textual embeddings as static targets, leading to suboptimal alignment. In this work, we identify the multi-granular heterophily in text-attributed graphs, including complete heterophily, partial heterophily, and latent homophily, which makes structure–text alignment particularly challenging due to mixed, noisy, and missing semantic correlations. To achieve flexible and bidirectional alignment, we propose GCL-OT, a novel graph contrastive learning framework with optimal transport, equipped with tailored mechanisms for each type of heterophily. Specifically, for partial heterophily, we design a RealSoftMax-based similarity estimator to emphasize key neighbor-word interactions while easing background noise. For complete heterophily, we introduce a prompt-based filter that adaptively excludes irrelevant noise during optimal transport alignment. Furthermore, we incorporate OT-guided soft supervision to uncover potential neighbors with similar semantics, enhancing the learning of latent homophily. Theoretical analysis shows that GCL-OT can improve the mutual information bound and Bayes error guarantees. Extensive experiments on nine benchmarks show that GCL-OT outperforms state-of-the-art methods, demonstrating its effectiveness and robustness.
\end{abstract}

\begin{links}
\link{Code, datasets and extended version}{github.com/users-01/GCL-OT}
\end{links}

\section{Introduction}\label{sec:intro}
\begin{figure}[t]
	\centering
	\begin{subfigure}[b]{0.47\linewidth}
		\centering
		\includegraphics[width=\linewidth]{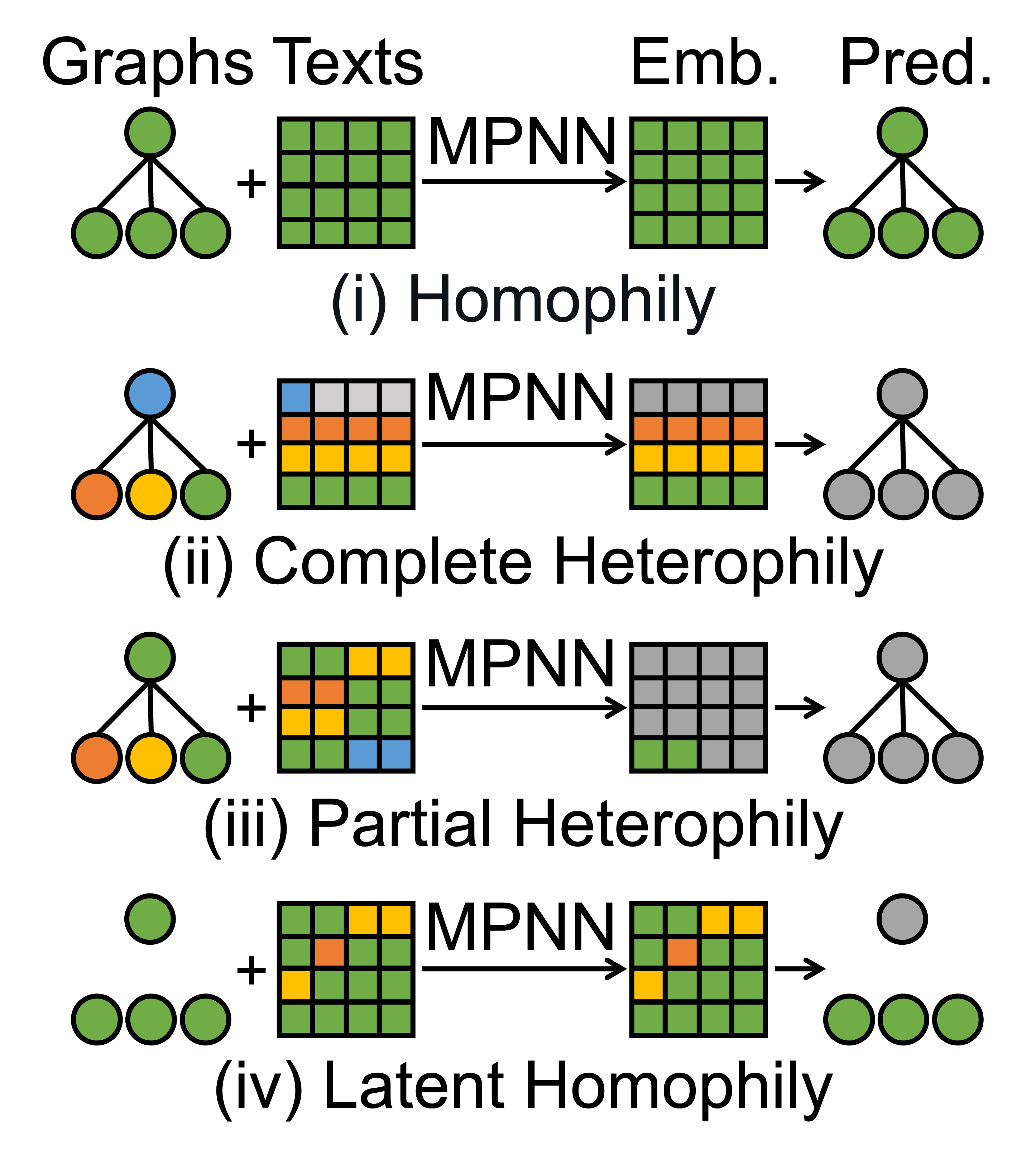}
		\caption{Examples}
		\label{fig:problem_example}
	\end{subfigure}
  \begin{subfigure}[b]{0.5\linewidth}
		\centering
		\includegraphics[width=\linewidth]{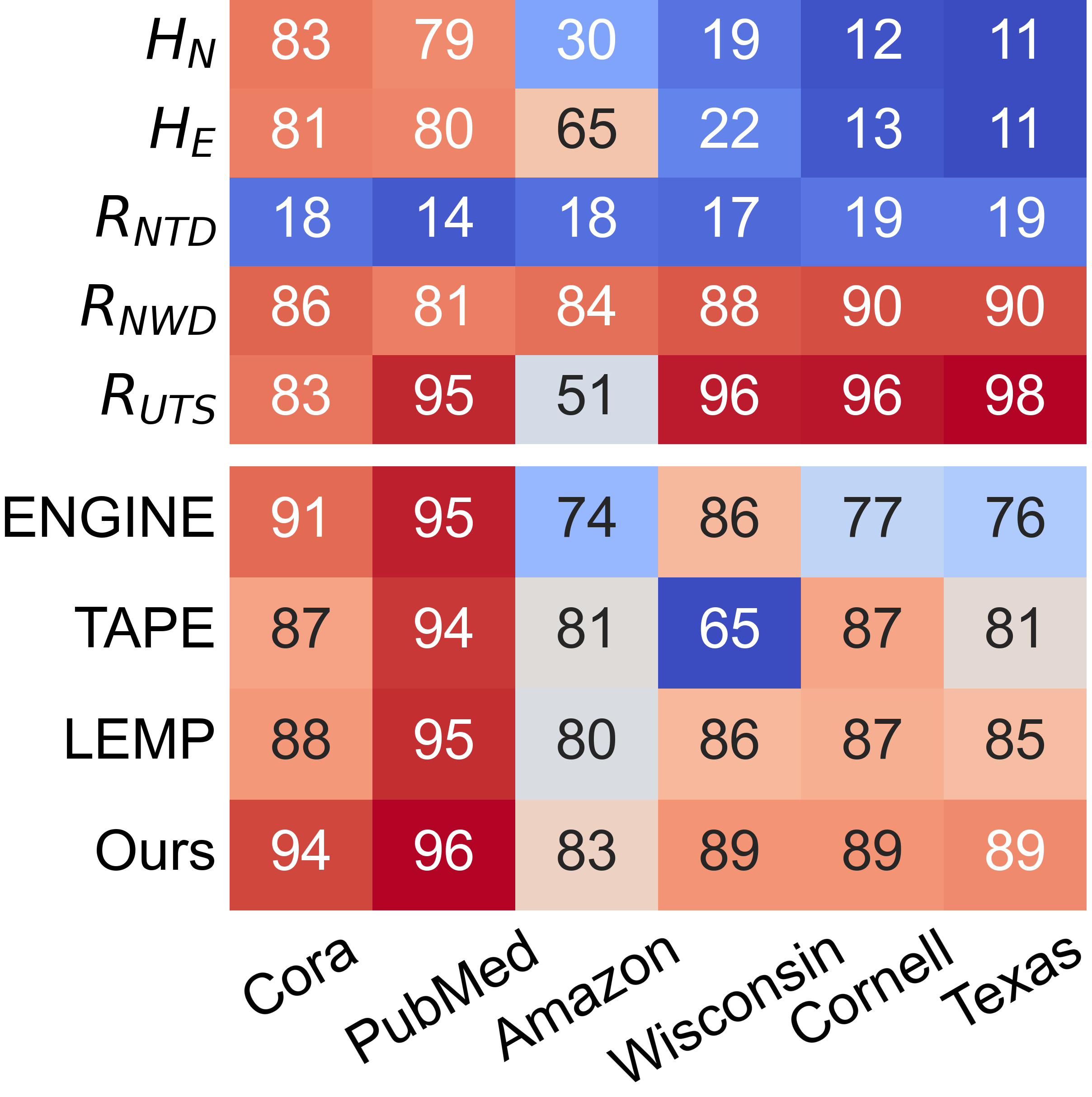}
		\caption{Empirical analysis}
		\label{fig:problem_statistics}
	\end{subfigure}
	\caption{Analysis of multi-granular heterophily in TAGs. Node colors denote categories. $H_N / H_E$: node/edge heterophily, $R_{\text{NTD}} / R_{\text{NWD}}$: neighbor token/sentence dissimilarity, $R_{\text{UTS}}$: similarity of unconnected nodes.}
	\label{fig:problem}
\end{figure}

Text-attributed graphs (TAGs) represent text entities as nodes and their relationships as edges, widely used across diverse real-world domains, such as academic citations, web hyperlinks, and e-commerce recommendations~\cite{yan2023tagbenchmark}. Recent advances in Large Language Models (LLMs) have shown substantial capability to capture the semantic richness, prompting researchers to combine Language Models (LMs) with Graph Neural Networks (GNNs) on TAGs for performance enhancement~\cite{he2024TAPE, pan2024Integrating, Liu2025GFM}. Among them, joint training strategies using Graph Contrastive Learning (GCL)~\cite{brannon2024congrat, Fang2024GAugLLM} enhance alignment between textual and structural representations by optimizing mutual information to distinguish positive and negative examples.

However, heterophilic TAGs are prevalent in reality due to the principle of opposites attracting, e.g., dating networks. Early GNN and GCL methods typically adopt the non-local neighbor extension and architectural refinement strategies~\cite{chen2024polygcl, wang2024HeterGCL}. Despite effectiveness, these approaches may be limited by the shallow embedding~\cite{Mikolov2013Distributed,zhang2010understanding}. Few methods study heterophily by combining deep semantic representations from LMs. Only LLM4HeG~\cite{wu2024LLM4HeG} and LEMP4HG~\cite{wang2025LEMP4HG} explore utilizing LLMs/LMs for heterophilic edge discrimination and reweighting, but face sub-optimal challenges with sparse or noisy data in the cascading pipelines~\cite{pan2024Integrating, Liu2025GFM}. 

In this paper, we identify that TAGs exhibit multi-granular heterophily from three node-centric perspectives grounded in the text-feature view, as illustrated in Figure~\ref{fig:problem_example}: 
(i) \emph{Partial heterophily} occurs when only part of a node’s text semantically aligns with its neighbors, and vice versa. Existing methods still struggle to capture subtle semantic mismatches~\cite{song2023orderedGNN}. 
(ii) \emph{Complete heterophily} appears when a node’s text has no relevance with its neighbors at all, typically caused by meaningless texts or connectivity (such as random co-purchases). Even state-of-the-art models can be misled by such signals~\cite{li2025HeTGB}. 
(iii) \emph{Latent homophily} denotes potential semantic neighbors that are disconnected, often due to missing or implicit links. While multi-hop propagation~\cite{zhu2020H2GCN} and potential neighbor discovery~\cite{Pei2020GeomGCN} exploit such signals, fully leveraging them remains open. 
As shown in Figure~\ref{fig:problem_statistics}, these patterns are prevalent in real-world TAGs and naturally induce many-to-many (N:N) alignment between structural and textual neighborhoods, whereas DGIs nearly N:1 summary pairing and InfoNCEs approximately 1:1 correspondence fail to capture this. Fortunately, optimal transport (OT) can fractionally allocate mass for soft N:N alignment when 1:1 is ambiguous, enabling mutual coverage via OT constraints to avoid one-sided greedy matches\cite{Xu2023GWF}. 
 
To address the multi-granular heterophily challenges in TAGs, we further propose \ours, a novel GCL framework with tailored mechanisms grounded in OT theory, aiming to hierarchically and softly align structural and textual representations. Specifically, for partial heterophily, the RealSoftMax-based similarity estimation can identify and emphasize the most relevant word–neighbor pairs bidirectionally. For complete heterophily, the global filter-prompt strategy is employed to mitigate the negative impact of irrelevant embeddings during alignment. For latent homophily, OT assignment serves as the auxiliary supervision within the GCL objective to discover hidden neighbors. We analyze \ours{} using mutual information (MI) theory, showing it can tighten the InfoNCE MI lower bound and reduce Bayes error in downstream node classification. Extensive experiments on nine TAG benchmarks demonstrate that \ours{} outperforms strong baselines, highlighting its robustness and potential in both homophilic and heterophilic settings. Our main contributions are summarized as follows: 
\begin{itemize}
\item To our knowledge, this is the first work to incorporate OT into GCL for heterophilic TAGs, enabling flexible bidirectional alignment of structural and textual views.
\item Three mechanisms tailored to multi-granular heterophily patterns: RealSoftMax-based similarity for partial heterophily, filter-prompt for complete heterophily, and OT-guided supervision for latent homophily. The theoretical analysis shows tighter MI guarantees.
\item Extensive experiments on nine benchmarks, in both homophilic and heterophilic scenarios, demonstrate the effectiveness and robustness of the proposed \ours{} compared with state-of-the-art methods.
\end{itemize}

\section{Related Work}
\label{sec:related}
\paragraph{Text-Attributed Graph Learning.} 
Representation learning on Text‑Attributed Graphs (TAGs) has attracted significant attention in graph machine learning. Traditional approaches typically couple shallow textual embeddings, such as bag-of-words or FastText, with GNNs~\cite{kipf2017GCN, hamilton2017SAGE, velivckovic2018GAT}. Recent efforts combine GNNs with LMs for richer textual encoding in cascading, iterative, and parallel manners. Cascading methods freeze GNN or LM as an auxiliary to enrich the other, while the two-stage, separately encoding often produce sub-optimal integration~\cite{he2024TAPE, wu2024LLM4HeG, wang2025LEMP4HG}. Iterative methods alternately optimize LMs and GNNs under the shared objective, improving label efficiency but potentially increasing computational costs~\cite{yang2021graphformers, zhao2023GLEM}. Parallel methods, mostly based on GCL, align LMs and GNNs in a shared space~\cite{zhu2024ENGINE, Fang2024GAugLLM}. Congrat~\cite{brannon2024congrat} and GraphGPT~\cite{tang2024graphgpt} align them by node-level InfoNCE objectives, which are intuitive but vulnerable to complex relations and noise. HASE-code~\cite{zhang2024hasecode} and G2P2~\cite{wen2023G2P2} extend the alignment granularity to subgraphs, but may risk over-smoothing due to high-order structures. Few approaches consider heterophily in TAGs. LLM4HeG~\cite{wu2024LLM4HeG} uses LLMs to label heterophilic edges from node-pair texts for reweighting. LEMP4HG~\cite{wang2025LEMP4HG} refines it via key-pair selection and attention integration. However, their two-stage pipelines make them sensitive to noisy text~\cite{pan2024Integrating, Liu2025GFM}.

\paragraph{Heterophilic Graph Learning.} 
Heterophily is generally considered a key challenge for GNN performance. Current researches typically fall into non-local neighbor extension and architectural refinement strategies~\cite{zheng2022graph, gong2024towards}. The first extends non-local neighbors with similar attributes through mixing high-order neighbors~\cite{abu2019mixhop, song2023orderedGNN} or discovering potential neighbors based on various distances~\cite {Pei2020GeomGCN}. The second refines GNN architecture by enhancing message aggregation from similar neighbors while minimizing the influence of dissimilar ones~\cite{bo2021beyond, zhu2021graph, liang2023predicting}, or mixing layers to capture information from different neighbor ranges~\cite{xu2018representation, zhu2020H2GCN}. Among them, recent GCL methods offer noteworthy advancements~\cite{chen2025m3pgcl}. PolyGCL~\cite{chen2024polygcl} designs and contrasts learnable spectral polynomial filters for varying homophily levels, while HeterGCL~\cite{wang2024HeterGCL} incorporates structural and semantic modules to utilize label-inconsistent signals effectively. However, these methods rely on static shallow embeddings and struggle to capture context-aware information and complex semantic relationships, limiting their effectiveness in leveraging text attributes~\cite{he2024TAPE}. 

\begin{figure*}[ht]
\centering
\includegraphics[width=0.8\textwidth]{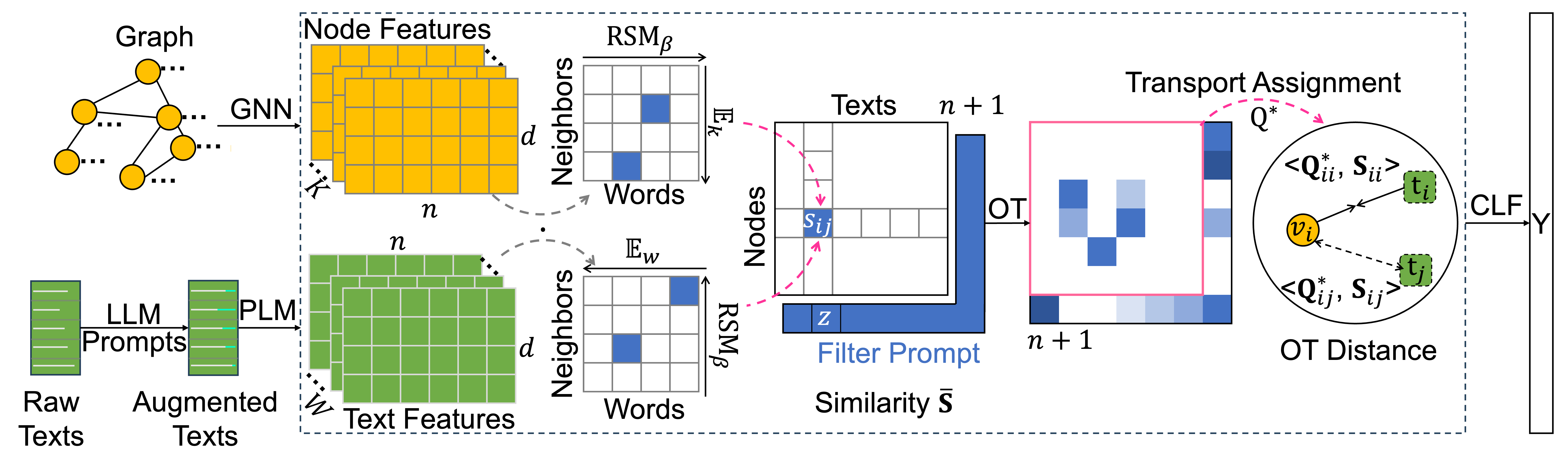}
\caption{Overview of \ours. Given a TAG, an LLM enriches node texts, a PLM encodes the enriched texts, and a GNN captures structure features. The text and structure views form a similarity matrix, where RealSoftMax highlights fine-grained interactions and the filter prompt suppresses coarse-grained noise. The contrastive module then aligns the two views and uncovers latent homophily. Finally, the fused embeddings drive node prediction.}
\label{fig:architecture}
\end{figure*}

\paragraph{Graph Contrastive Learning with OT} 
Optimal transport (OT)~\cite{Monge1781, Villani2009OT} is a mathematical framework for measuring distances between distributions, finding the most cost-efficient way to transform one distribution into another~\cite{Vincent2022OTGNN, lin2023multi, Xu2023GWF}. In Graph Contrastive Learning (GCL), traditional InfoNCE loss often relies on hard alignment with pairing positive and negative samples, which may introduce representation bias. To this end, researchers solve the problems caused by non-aligned contrast views or structural differences in the graph by introducing OT distance for soft alignment, subgraph generation, or cross-view comparison~\cite{wang2023galopa, deng2025THESAURUS}, thereby enhance node representation learning, which subsequently benefits downstream tasks such as node clustering ~\cite{Zhang2024CSOT, Wang2024PVCCL,deng2025THESAURUS}, node classification~\cite{xie2024SGEC, Amadou2025FOSSIL, Zhu2022RoSA, Liu2025STAR}, and node anomaly detection~\cite{Wang2023ACT}. Prior OT-based GCL methods (e.g., THESAURUS, FOSSIL) already treat OT as a soft-alignment remedy to InfoNCE’s hard pairing, but they operate on homogeneous graphs or prototype graphs.

\section{Method}\label{sec:methodology}\label{sec:OT_CL}

To address multi-granular heterophily in text-attributed graphs, we propose \ours, a graph contrastive learning framework with optimal transport designed to align structure and text representations gradually. The overall architecture of \ours{} is shown in Figure~\ref{fig:architecture}.

\subsection{Notations} 
Let $\mathcal{G}= (\mathcal{V},\mathcal{E},\mathcal{T}, \mathbf{y})$ denotes a text-attributed graph, where $\mathcal{V}=\{v_i\}_{i=1}^{N}$ is a set of $N$ nodes, $\mathcal{E}$ is a set of $M$ edges, $\mathcal{T}=\{t_i\}_{i=1}^{N}$ is text attributes of nodes, and $\mathbf{y}=[y_i]_{i=1}^{N}$ denotes node label vector drawn from a fixed set of categories. Given labeled nodes for training/validation, the goal is to predict labels for the remaining nodes.

\subsection{Prompt-Enhanced Multi-View Feature Encoding}

Heterophilic text-attributed graphs (TAGs) present a significant challenge, as node features and structural neighbors often belong to different classes. To overcome this inconsistency, we construct a multi-view encoding scheme that integrates both textual and structural signals.

Raw textual attributes are typically noisy and semantically ambiguous. To enrich their expressiveness, we augment each node's text using a task-specific prompt issued to a frozen large language model (LLM) following~\cite{he2024TAPE}. The original and prompted texts are concatenated to form an enhanced textual description $t_i^{\text{aug}}$ for each node $v_i$.

Then, we encode $t_i^{\text{aug}}$ with a lightweight PLM (e.g., DistilBERT~\cite{sanh2019distilbert}) to obtain token-level embeddings $\mathbf{H}^{\varpi}_i \in \mathbb{R}^{W \times d_t}$ and the sentence-level embedding $\mathbf{h}^t_i \in \mathbb{R}^{d_t}$. 
In parallel, we encode structures using a graph neural network (GNN). Node features are initialized and propagated through the GNN to produce structural embeddings $\mathbf{H}^{\zeta} \in \mathbb{R}^{N \times d_s}$, where $d_s$ is the structural dimension. For each node $v_i$, we derive a neighborhood-level embedding matrix $\mathbf{H}^{\mathcal{N}}_i \in \mathbb{R}^{K \times d_s}$ by aggregating and linear projecting the embeddings of its neighborhood $\mathcal{N}(v_i) = \{ v_k \mid e_{ik} \in \mathcal{E} \} \cup \{ v_i \}$, where $K = |\mathcal{N}(v_i)|$. These four views provide the foundation for contrastive objectives.

\subsection{GCL with OT for Hierarchical Alignment}

In partially heterophilic settings, simply averaging neighbor and word embeddings may weaken salient signals, while hard-max pooling is often extreme and unstable. To more balancedly emphasize meaningful interactions between neighbors and words, we introduce a soft maximum similarity estimation mechanism based on RealSoftMax.

Specifically, given the $k$-th neighbor embedding vector $\mathbf{h}^{\mathcal{N}}_{i,k}$ of node $v_i$, and the $w$-th word embedding vector $\mathbf{h}_{j,w}^{\varpi}$ of node $v_j$. The similarity between $v_i$ and $v_j$ is defined as
\begin{equation}
\begin{aligned}
 s_{ij} = \frac{1}{2}(
\mathbb{E}_{k}[\operatorname{RSM}_{\beta}(\{\mathbf{h}^{\mathcal{N}}_{ik} \cdot \mathbf{h}^{\varpi}_{jw}\}_{w=1}^{W})]\\
+
\mathbb{E}_{w}[\operatorname{RSM}_{\beta} (
\{\mathbf{h}^{\varpi}_{iw} \cdot \mathbf{h}^{\mathcal{N}}_{jk}\}_{k=1}^{K}
)]),
\end{aligned}
\label{eq:RealSoftMax}
\end{equation}
where $\mathbb{E}(\cdot)$ is the mean operator, $\operatorname{RSM}_{\beta}(\{x_\ell\})=\beta\log\sum_\ell\exp(x_\ell/\beta)$ smoothly interpolates between mean ($\beta\to \infty$) and maximum ($\beta \to 0$). The first term highlights the most relevant words for each neighbor, the second does the opposite, jointly emphasizing informative cross-view interactions and downweighting background noise.

To address the challenge of complete heterophily, we introduce a prompt-based filtering mechanism that reduces noise that can not align with others. Specifically, the global similarity matrix $\hat{\mathcal{S}}$ between structural embedding $\mathbf{H}^{\zeta}$ and textual embedding $\mathbf{H}^{t}$ is merged into $\mathbf{S}$, and then $\mathbf{S}$ is expanded with additional row and column prompt vectors,
\begin{equation}
\bar{\mathbf{S}} =
\begin{bmatrix}
\mathbf{S} & \mathbf{z} \\
\mathbf{z}^\top & z_{N+1}
\end{bmatrix}
\in \mathbb{R}^{(N+1)\times(N+1)},
\label{eq:s_hat_prompt}
\end{equation}
where $\mathbf{z}$ is a learnable vector. Embeddings whose maximum similarity falls below the corresponding $\mathbf{z}$ are instead aligned with the prompt vector during the subsequent alignment.

Based on $\bar{\mathbf{S}}$, we adopt the OT distance as the similarity measure for robust and efficient alignment. Let $\mathbf{Q}\in \mathbb{R}_{+}^{(N+1)\times (N+1)}$ denote the transport assignment of $\bar{\mathbf{S}}$, where $q_{ij}$ is the corresponding probabilities. OT aims to establish a flexible alignment by maximizing $\langle \mathbf{Q}, \bar{\mathbf{S}}\rangle=\operatorname{tr}(\mathbf{Q}^\top\bar{\mathbf{S}})$. The optimization problem can be defined as
\begin{equation}
 \begin{aligned}
& \max _{\mathbf{Q} \in \mathcal{Q}} \langle\mathbf{Q}, \bar{\mathbf{S}}\rangle +\varepsilon \operatorname{H}(\mathbf{Q})
 \\
& \text {s.t. } 
\mathcal{Q}=\left\{\mathbf{Q}
\mid \mathbf{Q} \mathbf{1}_{N+1}=\boldsymbol{\mu}_1, \mathbf{Q}^{\top} \mathbf{1}_{N+1}= \boldsymbol{\mu}_2 \right\},
\end{aligned}
\label{eq:ot}
\end{equation}
where $\mathbf{1}_{N+1} \in \mathbb{R}^{N+1}$ is an all-one vector, entropy term $\operatorname{H}(\mathbf{Q}) =-\sum_{ij} q_{ij} \log q_{ij}$ is a smooth convex regular, $\varepsilon$ is weight of entropy, $\boldsymbol{\mu}_1\in \mathbb{R}^{N+1}$ and $\boldsymbol{\mu}_2\in \mathbb{R}^{N+1}$ are relative importance of structural and textual embeddings. To avoid bias, let $\boldsymbol{\mu}_1$ and $\boldsymbol{\mu}_2$ follow the uniform distribution $\frac{\mathbf{1}_{N+1}}{N+1}$.

As shown in Equation~\eqref{eq:ot}, OT can realign each node's structure or text with multiple related texts or structures based on similarity, effectively resolving potential issues between structures and texts. Let the kernel matrix $\mathbf{K} = \exp(\bar{\mathbf{S}}/\varepsilon)$, and we approximate $\mathbf{K}$ with a non-negative low-rank factorization
 \begin{equation}
\begin{aligned}
\mathbf{K} \approx	\mathbf{U} \operatorname{diag}(\Upsilon) \mathbf{V}^{\top},
\end{aligned}
\label{eq:kernel_lr}
\end{equation}
where $\mathbf{U}\in\mathbb{R}_{+}^{(N+1)\times r}$ and $\mathbf{V}\in\mathbb{R}_{+}^{(N+1)\times r}$ are non-negative basis matrices, $\operatorname{diag}(\Upsilon) \in \mathbb{R}^{r \times r}$ turns the weight vector into a diagonal matrix, and target rank $r \leq N$. The optimal $\mathbf{Q}^*$ of Equation~\eqref{eq:ot} has a simple normalized exponential matrix solution by LRSinkhorn~\cite{Scetbon2021lrsinkhorn}:
\begin{equation}
\begin{aligned}
\mathbf{Q}^* = [\operatorname{Diag}(\boldsymbol{\kappa}_1) \mathbf{K} \operatorname{Diag}(\boldsymbol{\kappa}_2)]_{1: N,1:N},
\end{aligned}
\label{eq:lrsinkhorn}
\end{equation}
with iteratively updated
\begin{equation}
\begin{aligned}
\left\{\begin{matrix}
\boldsymbol{\kappa}_1 \leftarrow \boldsymbol{\mu}_1 . / ( \mathbf{K} \cdot \boldsymbol{\kappa}_2 ), \\
\boldsymbol{\kappa}_2 \leftarrow \boldsymbol{\mu}_2 . / ( \mathbf{K}^{\top} \cdot \boldsymbol{\kappa}_1),
\end{matrix}\right. 
\end{aligned}
\label{eq:update}
\end{equation}
where $\boldsymbol{\kappa}_1 \in \mathbb{R}^N$, $\boldsymbol{\kappa}_2 \in \mathbb{R}^N$ are the non-negative left and right scaling vectors.

By utilizing OT distance as the similarity criterion, and let 
\begin{equation}
d_{ij}= \exp(q^{*}_{ij} \bar{s}_{ij}/\tau),
\end{equation}
where $q^*_{ij}$ is the corresponding transport assignment of $\bar{s}_{ij}$, and $\tau$ is the temperature parameter. 
To encourage strong alignment of matching pairs while repelling mismatches, we minimize the bi-directional contrastive loss $\mathcal{L}_{\text{MHA}}$ as
\begin{equation}
\begin{aligned}
\mathcal{L}_{\text{MHA}} = -\mathbb{E}_{i}\left[\log\frac{{d_{ii}}}{\sum_{j=1}^{r}d_{ij}}+\log\frac{d_{ii}}{\sum_{j=1}^r d_{ji}}\right],
\end{aligned}
\label{eq:loss_align}
\end{equation}
where we retain $r$ samples with the largest scores as negatives. For irrelevant inputs, the OT plan $ q^{*}_{ij}$ distributes mass more broadly, resulting in a lower alignment score $s_{ij}$ and a decreased affinity $d_{ij}$. Thus, minimizing $\mathcal{L}_{\textbf{MHA}} $ penalizes them by increasing distance in embedding space, enhancing the contrastive signal.

\subsection{OT-Guided Latent Homophily Mining}\label{sec:Latent_Neighbor_Exploration}

Randomly sampled negative samples and one-hot labels may mistakenly penalize semantically similar but unconnected potential neighbors as negative samples. We leverage the OT assignment matrix as auxiliary supervision to reduce the distances between potential neighbors.

Specifically, given the global similarity matrix $\hat{\mathbf{S}}$ between structural embeddings $\mathbf{H}^{\zeta}$ and textual embeddings $\mathbf{H}^{t}$, the OT problem can be formulated and solved as illustrated in Equation~\eqref{eq:ot}–\eqref{eq:update}, then it generates the assignment $\hat{\mathbf{Q}}^*$.

Considering the diagonal matrix $\mathbf{I}$ encoding self-positives, and $\hat{\mathbf{Q}}^*$ encoding latent positives. The contrastive target $\mathbf{P}$ can be defined as their combination:
\begin{equation}
\begin{aligned}
\mathbf{P} = \mathbf{I} + \hat{\mathbf{Q}}^*,
\end{aligned}
\end{equation}
where $\mathbf{P}$ is normalized to obtain a valid probability distribution for each anchor. Based on the soft target $\mathbf{P}$, the Latent Homophily Mining (LHM) loss can be defined as
\begin{equation}
\begin{aligned}
\mathcal{L}_{\textbf{LHM}}
= -\frac{1}{N} \sum_{i=1}^N \sum_{j=1}^r p_{ij}
(\log \hat{k}_{ij}^{'} + \log \hat{k}^{''}_{ij})
\end{aligned}
\label{eq:loss_LHM}
\end{equation}
where $\hat{k}_{ij}^{'} = \operatorname{softmax}_j(\hat{s}_{ij}/\tau)$ denotes the row-wise softmax over index $j$, $\hat{k}_{ij}^{''} = \operatorname{softmax}_i(\hat{s}_{ij}/\tau)$ denotes denotes the row-wise softmax over index $i$. 
By minimizing $\mathcal{L}_{\textbf{LHM}}$, \ours{} can implicitly increase the mutual attraction among similar embeddings, thereby strengthening the contrastive learning objective. 

\subsection{Optimization Objective}\label{sec:optimize_object}
To achieve both effective alignment between textual and structural information and strong discriminative power for node classification, \ours{} is trained with a composite objective that couples a contrastive component with the standard node-classification term,
\begin{equation}
\begin{aligned}
\mathcal{L}=\mathcal{L}_{\text{NC}}+\lambda \mathcal{L}_{\text{GCL-OT}},
\end{aligned}
\label{eq:loss}
\end{equation}
where $\mathcal{L}_{\text{NC}}$ is the cross-entropy loss for node classification, $\lambda$ balances contributions of the contrastive term. 

\subsection{Theoretical Analysis}\label{sec:theory}
In this section, we explain the effectiveness of \ours{} from the view of Mutual Information (MI) theory. 

\begin{proposition}\label{theory:proposition_MHA}
Let $\mathcal{L}_{\text{InfoNCE}}$ denote the standard InfoNCE loss between structural embeddings $\mathbf{H}^{\zeta}$ and textual embeddings $\mathbf{H}^{t}$. According to the standard InfoNCE MI lower bound~\cite{oord2019infonce}, $\mathcal{L}_{\text{LHM}}$ provides a tighter variational lower bound than InfoNCE
\begin{equation}
\operatorname{MI}(\mathbf{H}^{\zeta}, \mathbf{H}^{t}) \geq \log N - \mathcal{L}_{\text{MHA}} \geq \log N - \mathcal{L}_{\text{InfoNCE}}.
\end{equation}
\end{proposition}

\begin{proposition}\label{theory:proposition_lhm}
According to the standard InfoNCE MI lower bound~\cite{oord2019infonce}, the latent-homophily objective $\mathcal{L}_{\text{LHM}}$ also satisfies
\begin{equation}
\operatorname{MI}(\mathbf{H}^{\zeta}, \mathbf{H}^{t}) \geq \log N - \mathcal{L}_{\text{LHM}} \geq \log N - \mathcal{L}_{\text{InfoNCE}}.
\end{equation}
\end{proposition}

Propositions~\ref{theory:proposition_MHA} and~\ref{theory:proposition_lhm} claim that minimizing $\mathcal{L}_{\text{MHA}}$ and $\mathcal{L}_{\text{LHM}}$ for optimization maximizes a variational lower bound on mutual information than $\mathcal{L}_{\text{InfoNCE}}$. Based on these propositions, we can derive Theorem~\ref{theory:theorem}.

\begin{theorem}\label{theory:theorem}
Let $f^{\zeta}(\cdot)$ and $f^{t}(\cdot)$ denote deterministic encoder functions for structural and textual information, respectively. For each node $v_i$, let $\mathbf{x}_i^{(k)}=\{\mathbf{x}_j\}_{j \in \mathcal{N}(\mathbf{x}^{(k)}, i)}$ denote its $k$-hop neighborhood that collectively maps to its high-level features, $\mathbf{h}_i^{\zeta}=f^{\zeta}(\mathbf{x}_i^{(k)})$, $\mathbf{h}_i^{t}=f^{t}(\mathbf{x}_i^{(k)})$. Let $\mathbf{h}_i=\operatorname{Linear}(\mathbf{h}_i^{\zeta}, \mathbf{h}_i^{t})$ denotes the final embedding. Assume that the weight matrix of the $\operatorname{Linear}(\cdot,\cdot)$ has full column rank. Then minimising loss in Equation~\eqref{eq:loss} maximizes the variational lower bounds on $\operatorname{MI}(\mathbf{H}^{\zeta},\mathbf{H}^{t})$ and thus $\operatorname{MI}(\mathbf{x}_i^{(k)},\mathbf{h}_i)$.
\end{theorem}

\paragraph{Remark.}\label{theory:remark}
For each view $i \in \{\zeta,t\}$, let $ P_{\text{pos}}^{i}$ and $P_{\text{neg}}^{i}$ denote the positive and negative pair distributions, $\operatorname{JS}^{\zeta}(P_{\text{pos}}^{\zeta},P_{\text{neg}}^{\zeta})$ and $\operatorname{JS}^{t}(P_{\text{pos}}^{t},P_{\text{neg}}^{t})$ denote the corresponding Jensen-Shannon divergences (JSD). Assume that under the standard one positive and many negative construction and the optimal discriminators, we can rewrite $\mathcal{L}_{\text{GCL-OT}} \geq 4 \log{r} -2 (\operatorname{JS}^{\zeta}+\operatorname{JS}^{t})$, which proves that minimizing $\mathcal{L}_{\text{GCL-OT}}$ not only maximizes MI, but also encourages a larger JSD between positive and negative distributions in both structural and textual views.

\paragraph{Connection with downstream tasks.} 
Optimizing objective in Equation~\eqref{eq:loss}, \ours{} provides a tighter upper bound on the downstream Bayes error~\cite{tsai2021sslmultiview, xiao2022decoupled} compared to $\mathcal{L}_{\textbf{InfoNCE}}$ and $\mathcal{L}_{\textbf{NC}}$. This suggests that downstream tasks can benefit from the learned representations obtained through our loss.

\subsection{Time Complexity Analysis}\label{sec:complexity}
Training in \ours{} mainly consists of feature encoding and contrastive learning. The GNN encoder (GCN) costs $\mathcal{O}(|E|D)$, the PLM encoder costs $\mathcal{O}(N W^{2} D)$. RealSoftMax similarity over $K$ neighbors and $W$ tokens costs $\mathcal{O}(N K W D)$, with $K \ll N$. Projecting both embeddings onto $r$-rank bases and keeping $r$ similarities per row costs $\mathcal{O}(N r D)$. The rank-$r$ LR Sinkhorn costs $\mathcal{O}(N r + r^{2})$. The contrastive losses add $2 \mathcal{O}(N r)$. Consequently, the overall per-epoch complexity is $\mathcal{O}(|E|D + N W^{2} D + N r D + Nr)$. In typical benchmarks, the $\mathcal{O}(N W^{2} D)$ term dominates, so \ours{} scales nearly linearly with the number of nodes and feature dimension, and quadratically with text length.

\begin{table*}
\centering
\small
\setlength{\tabcolsep}{1mm}
\begin{tabular}{@{}lccccccccc@{}}
\toprule
Methods & Cora & PubMed & Products & ArXiv & Amazon & ArXiv23 & Wisconsin & Cornell & Texas \\
\midrule
$H_\textbf{N}$ & 82.52 & 79.24 & 78.97 & 63.53 & 37.57 & 29.66 & 18.68 & 12.12 & 10.68 \\
\midrule
MLP & 63.88$\pm$2.1 & 86.35$\pm$0.3 & 61.06$\pm$0.1 & 53.36$\pm$0.4 & 47.87$\pm$0.9 & 62.02$\pm$0.6 & 85.85$\pm$6.8 & 72.05$\pm$7.1 & 80.26$\pm$7.3 \\
DistilBERT & 76.06$\pm$3.8 & 94.94$\pm$0.5 & 72.97$\pm$0.2 & 73.61$\pm$0.0 & 65.42$\pm$0.7 & 73.58$\pm$0.1 & 86.79$\pm$5.5 & 79.49$\pm$4.5 & 62.50$\pm$14.5 \\
GPT3.5 & 67.69 & 93.42 & 74.40 & 73.50 & 37.71 & 73.56 & 62.26 & 69.23 & 65.78 \\
GCN & 89.11$\pm$0.2 & 85.33$\pm$1.0 & 75.64$\pm$0.2 & 71.82$\pm$0.3 & 45.14$\pm$0.6 & 63.41$\pm$0.6 & 46.98$\pm$9.3 & 44.36$\pm$5.6 & 54.21$\pm$7.9 \\
GAT & 88.24$\pm$0.9 & 88.75$\pm$0.9 & 79.45$\pm$0.6 & 73.95$\pm$0.1 & 43.55$\pm$0.4 & 64.53$\pm$0.5 & 44.91$\pm$9.9 & 46.67$\pm$12.3 & 51.32$\pm$9.9 \\
SAGE & 88.24$\pm$0.1 & 88.81$\pm$0.0 & 78.29$\pm$0.2 & 71.71$\pm$0.2 & 45.00$\pm$0.6 & 64.30$\pm$0.4 & 81.13$\pm$8.4 & 69.49$\pm$7.8 & 78.16$\pm$7.3 \\
H2GCN & 87.81$\pm$1.4 & 89.59$\pm$0.3 & 48.62$\pm$0.1 & 72.50$\pm$0.2 & 41.64$\pm$0.7 & 77.61$\pm$2.9 & 87.74$\pm$6.3 & 83.68$\pm$5.1 & 84.21$\pm$5.4 \\
FAGCN & 88.85$\pm$1.4 & 89.98$\pm$0.5 & 67.94$\pm$0.2 & 71.83$\pm$0.2 & 44.20$\pm$0.7 & 74.46$\pm$0.4 & 49.43$\pm$4.2 & 49.74$\pm$9.3 & 53.95$\pm$6.5 \\
GIANT-BERT & 85.52$\pm$0.7 & 85.02$\pm$0.5 & 74.06$\pm$0.4 & 74.26$\pm$0.2 & 52.31$\pm$0.6 & 72.18$\pm$0.3 & 86.38$\pm$3.6 & 75.74$\pm$3.6 & 77.87$\pm$2.8 \\
GLEM-DeBERTa & 85.60$\pm$0.1 & 94.71$\pm$0.2 & 73.77$\pm$0.1 & 74.69$\pm$0.3 & 50.18$\pm$0.8 & 78.58$\pm$0.1 & 87.14$\pm$4.1 & 84.76$\pm$4.6 & 86.76$\pm$5.2 \\
ENGINE-LLAMA & 91.48$\pm$0.3 & 95.24$\pm$0.4 & 80.05$\pm$0.5 & 76.02$\pm$0.3 & 54.60$\pm$0.9 & 79.76$\pm$0.1 & 85.50$\pm$4.0 & 77.36$\pm$4.5 & 75.68$\pm$5.0 \\
TAPE-GCN & 87.41$\pm$1.6 & 94.31$\pm$0.4 & 79.96$\pm$0.4 & 75.20$\pm$0.0 & 48.14$\pm$4.1 & 80.80$\pm$2.2 & 61.79$\pm$14.3 & 87.32$\pm$1.8 & 81.36$\pm$0.4 \\
TAPE-RevGAT & 92.80$\pm$2.8 & 96.04$\pm$0.5 & 79.76$\pm$0.1 & 77.50$\pm$0.1 & 47.22$\pm$1.0 & 79.95$\pm$0.6 & 87.77$\pm$7.0 & \underline{ 88.46$\pm$5.3} & 85.90$\pm$3.3 \\
TAPE-SAGE & 92.90$\pm$3.1 & 94.80$\pm$0.4 & 81.37$\pm$0.4 & 76.72$\pm$0.0 & 46.39$\pm$2.4 & 80.23$\pm$0.3 & \underline{ 88.89$\pm$3.4} & 87.18$\pm$4.7 & 85.26$\pm$6.4 \\
SimTeG-e5 & 88.04$\pm$1.4 & 94.84$\pm$0.8 & 74.51$\pm$1.5 & 75.29$\pm$0.2 & 48.95$\pm$1.2 & 79.51$\pm$0.5 & 87.12$\pm$2.8 & 85.04$\pm$3.9 & 84.36$\pm$3.0 \\
LEMP-TAPE & 88.26$\pm$1.2 & 94.85$\pm$0.3 & -- & -- & 46.75$\pm$2.1 & 80.03$\pm$0.2 & 85.65$\pm$4.6 & 86.54$\pm$3.2 & 85.09$\pm$5.3 \\
\midrule
Ours-GCN & \underline{ 93.54$\pm$1.3} & 96.08$\pm$0.9 & 81.50$\pm$0.1 & \textbf{78.15$\pm$0.1} & \textbf{66.40$\pm$0.4} & \underline{ 82.60$\pm$2.2} & 88.68$\pm$7.1 & \textbf{88.64$\pm$6.0} & \underline{ 89.47$\pm$3.4} \\
Ours-GAT & 92.98$\pm$1.6 & \underline{ 96.14$\pm$1.2} & \textbf{82.24$\pm$0.4} & 77.50$\pm$0.1 & 66.12$\pm$0.3 & 81.63$\pm$2.6 & 81.13$\pm$12.3 & 87.69$\pm$5.6 & 88.16$\pm$5.6 \\
Ours-SAGE & \textbf{93.73$\pm$1.9} & \textbf{96.62$\pm$0.6} & \underline{ 81.73$\pm$0.5} & \underline{ 78.13$\pm$0.2} & \underline{ 66.28$\pm$0.5} & \textbf{82.72$\pm$2.8} & \textbf{89.26$\pm$4.9} & 88.21$\pm$6.0 & \textbf{90.01$\pm$4.4}\\
\bottomrule
\end{tabular}
\caption{Mean node classification accuracy (\%). The best and second-best results are in \textbf{bold} and \underline{underlined}, respectively. -- denotes unavailable results. $H_N$ is the node-level homophily metric.} 
\label{tab:nc}
\end{table*}

\begin{table}[htbp]
\centering
\begin{tabular}{@{}lccc@{}}
\toprule
Methods & Wisconsin & Cornell & Texas \\ \midrule
DGI & 55.21$\pm$1.0 & 45.33$\pm$6.1 & 58.5$\pm$3.0 \\
GRACE & 47.11$\pm$3.5 & 41.48$\pm$3.8 & 53.55$\pm$3.5 \\
Congrat & 63.00$\pm$4.9 & 57.90$\pm$4.1 & 62.60$\pm$4.3 \\
PolyGCL & 70.24±4.3 &62.45±5.5 & 66.77±5.5\\
HeterGCL & \underline{71.93±4.2} & \underline{64.02±4.8} & \underline{72.79±2.5}\\
Ours-GCN & \textbf{72.83$\pm$7.1} & \textbf{64.10$\pm$4.4} & \textbf{73.68$\pm$3.4} \\ 
\bottomrule
\end{tabular}
\caption{Mean node classification accuracy (\%) under the unsupervised setting.}
\label{tab:unsupervised_nc}
\end{table}

\section{Experiments}\label{sec:experiments}

\subsection{Experimental Settings}\label{sec:experiments-setup}
To evaluate the performance of \ours{}, the node classification tasks are conducted on 9 real-world datasets across different homophily degrees. Cora, PubMed~\cite{sen2008collective_pubmed}, ArXiv, and Products~\cite{hu2020ogb} generally regarded as homophilic, whereas Amazon~\cite{platonov2023critical}, ArXiv23~\cite{he2024TAPE}, Wisconsin, Cornell, and Texas~\cite{Pei2020GeomGCN} fall into heterophilic category. Original textual information for heterophilic graphs was collected from publicly available sources~\cite{leskovec2007dynamics,craven1998learning,wu2024LLM4HeG}.

Experiments are conducted on one NVIDIA 4090 GPU and three NVIDIA 3080 GPUs. Text augmentation uses GPT-3.5 following~\cite{he2024TAPE}. The text encoder employs a partially frozen DistilBERT~\cite{sanh2019distilbert} model with six layers. Graph encoder adopts GCN~\cite{kipf2017GCN}, GAT~\cite{velivckovic2017graph_gat}, and GraphSAGE~\cite{hamilton2017SAGE}, each configured with two layers. GAT utilizes eight attention heads, and GraphSAGE samples all neighbors. The classifier adopts a three-layer MLP with ReLU. Dimension of Text and Graph encoder is set to 768, with learning rates of 0.01 and 2e-5, weight decay at 0.0005, dropout from 0 to 0.8, and a batch size of 512. ArXiv and Products follow the standard splits from~\cite{hu2020ogb}, the other datasets use random splits (60/20/20) following~\cite{he2024TAPE}. Baseline hyperparameters follow prior work. All experiments report the average accuracy and standard deviation over 10 random seeds. 

\subsection{Experimental Results}\label{subsec:experiments-PerformanceComparison}

\begin{figure}[tb]
\centering
\includegraphics[width=0.75\linewidth]{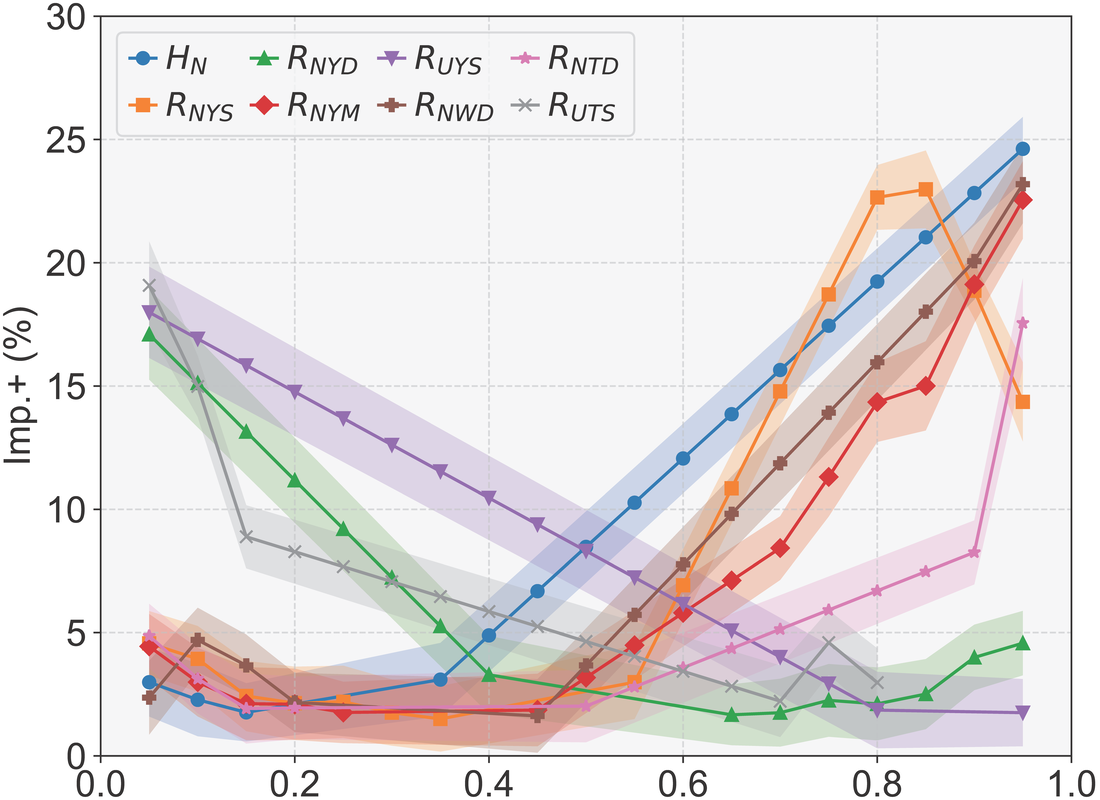}
\caption{Improvements over InfoNCE across various heterophily metrics on Cora.}
\label{fig:metric_improv}
\end{figure}

\paragraph{Case Study.}\label{subsubsec:experiments-Empirical}
We first study the improvements over InfoNCE across various heterophily metrics by perturbing edges and texts. As shown in Figure~\ref{fig:metric_improv}, \ours{} achieves substantial gains over the InfoNCE variant, obtained by replacing $\mathcal{L}_{\text{GCL-OT}}$ with $\mathcal{L}_{\text{InfoNCE}}$, with significant improvements under mixed-label neighborhoods and strong semantic heterophily. These trends indicate that \ours{} can handle label mixing and semantic disagreement within neighborhoods, while the abundance of purely heterophilic neighborhoods and unlinked similar pairs may limit its benefits.

\begin{figure}[hbp]
\centering
\begin{subfigure}[b]{0.3\linewidth}
\centering
\includegraphics[width=\linewidth]{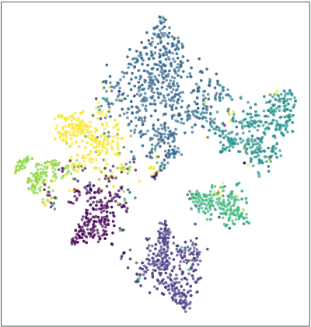}
\caption{GCN}
\label{fig:cora_gcn}
\end{subfigure}
\begin{subfigure}[b]{0.3\linewidth}
\centering
\includegraphics[width=\linewidth]{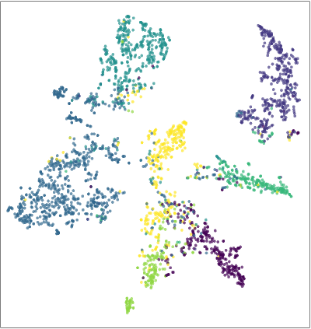}
\caption{TAPE-GCN-E}
\label{fig:cora_tape}
\end{subfigure}
\begin{subfigure}[b]{0.3\linewidth}
\centering
\includegraphics[width=\linewidth]{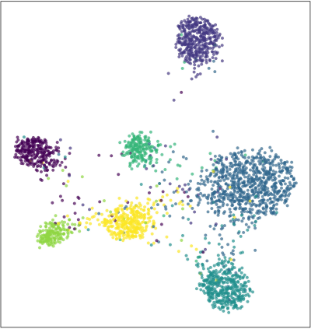}
\caption{\ours{}-GCN}
\label{fig:cora_our}
\end{subfigure}
\\
\begin{subfigure}[b]{0.3\linewidth}
\centering
\includegraphics[width=\linewidth]{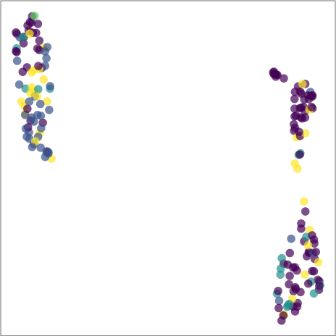}
\caption{GCN}
\label{fig:texas_gcn}
\end{subfigure}
\begin{subfigure}[b]{0.3\linewidth}
\centering
\includegraphics[width=\linewidth]{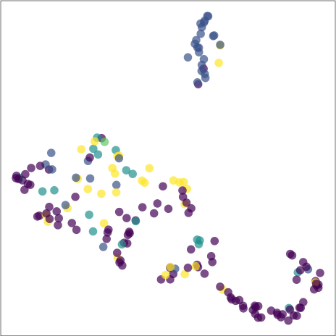}
\caption{TAPE-GCN-E}
\label{fig:texas_tape}
\end{subfigure}
\begin{subfigure}[b]{0.3\linewidth}
\centering
\includegraphics[width=\linewidth]{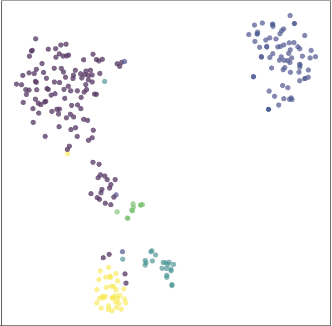}
\caption{\ours{}-GCN}
\label{fig:texas_our}
\end{subfigure}
\caption{T-SNE visualization of node representations learned by different models on Cora (a–c) and Texas (d–f), with colors indicating ground-truth class labels.}
\label{fig:tsne}
\end{figure}

\paragraph{Results on Node Classification.}
Table~\ref{tab:nc} shows that \ours{} substantially outperforms MLP, LM-only, and classical GNN baselines under the supervised setting, highlighting the benefit of jointly modeling text and structure. It also surpasses TAG methods and heterophilic GNNs, indicating that \ours{} can effectively mitigate heterophily rather than overfitting to homophilic patterns. In the unsupervised setting, we train \ours{} on raw node texts with no labels, then freeze the model and train a one-layer linear classifier on 60\%/20\%/20\% splits of the embeddings and labels. As shown in Table~\ref{tab:unsupervised_nc}, \ours{} outperforms classical and heterophilic GCL methods, demonstrating that \ours{} can learn discriminative representations even without supervision

\paragraph{Visualization of Learned Representations.}
We further conduct t-SNE visualization on learned node embeddings. The results on Cora, Actor, and Texas datasets are shown in Figure~\ref{fig:tsne}. It can be observed that compared to baselines, \ours{} produces more semantically coherent and better-separated clusters.

\begin{figure*}
\centering
\begin{subfigure}[b]{0.3\linewidth}
\centering
\includegraphics[width=\linewidth]{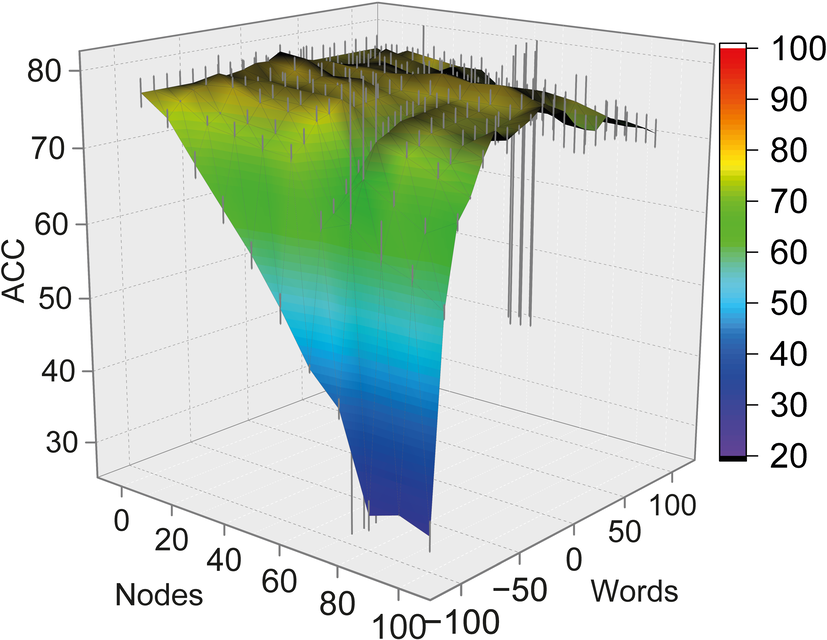}
\caption{DistilBERT on text perturbation (\%) }
\label{fig:perturb_node_lm}
\end{subfigure}
\begin{subfigure}[b]{0.3\linewidth}
\centering
\includegraphics[width=\linewidth]{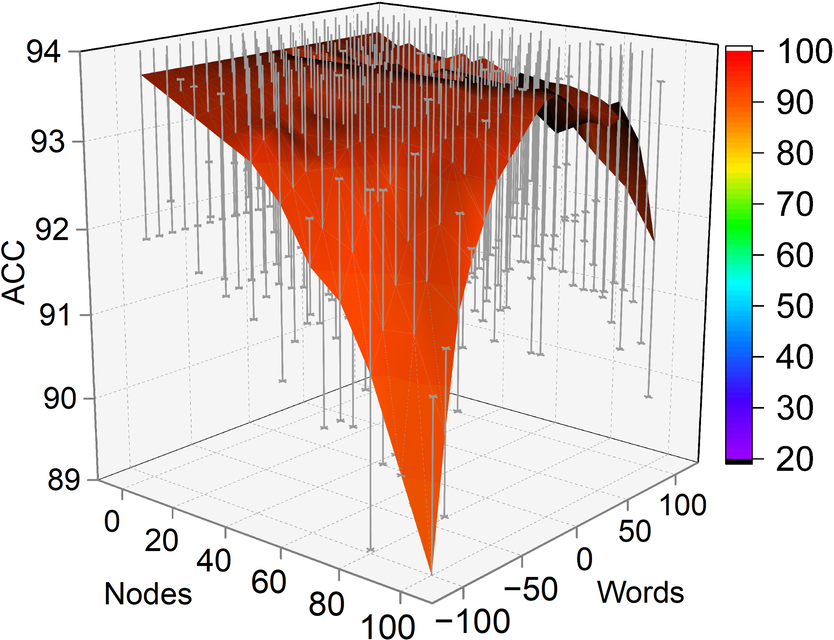}
\caption{\ours{} on text perturbation (\%)}
\label{fig:perturb_node_ours}
\end{subfigure}
\begin{subfigure}[b]{0.3\linewidth}
\centering
\includegraphics[width=\linewidth]{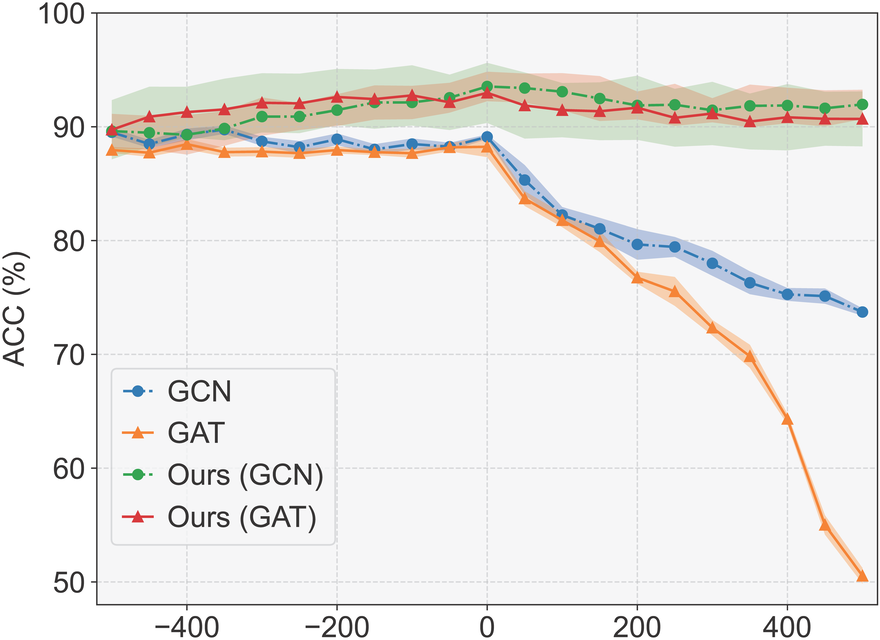}
\caption{Models on edge perturbation}
\label{fig:perturb_edge}
\end{subfigure}
\caption{Evaluation of model robustness under text and edge perturbations on Cora.}
\label{fig:robust}
\end{figure*}

\begin{figure*}[htbp]
\centering
\begin{subfigure}[b]{0.3\linewidth}
\centering
\includegraphics[width=\linewidth]{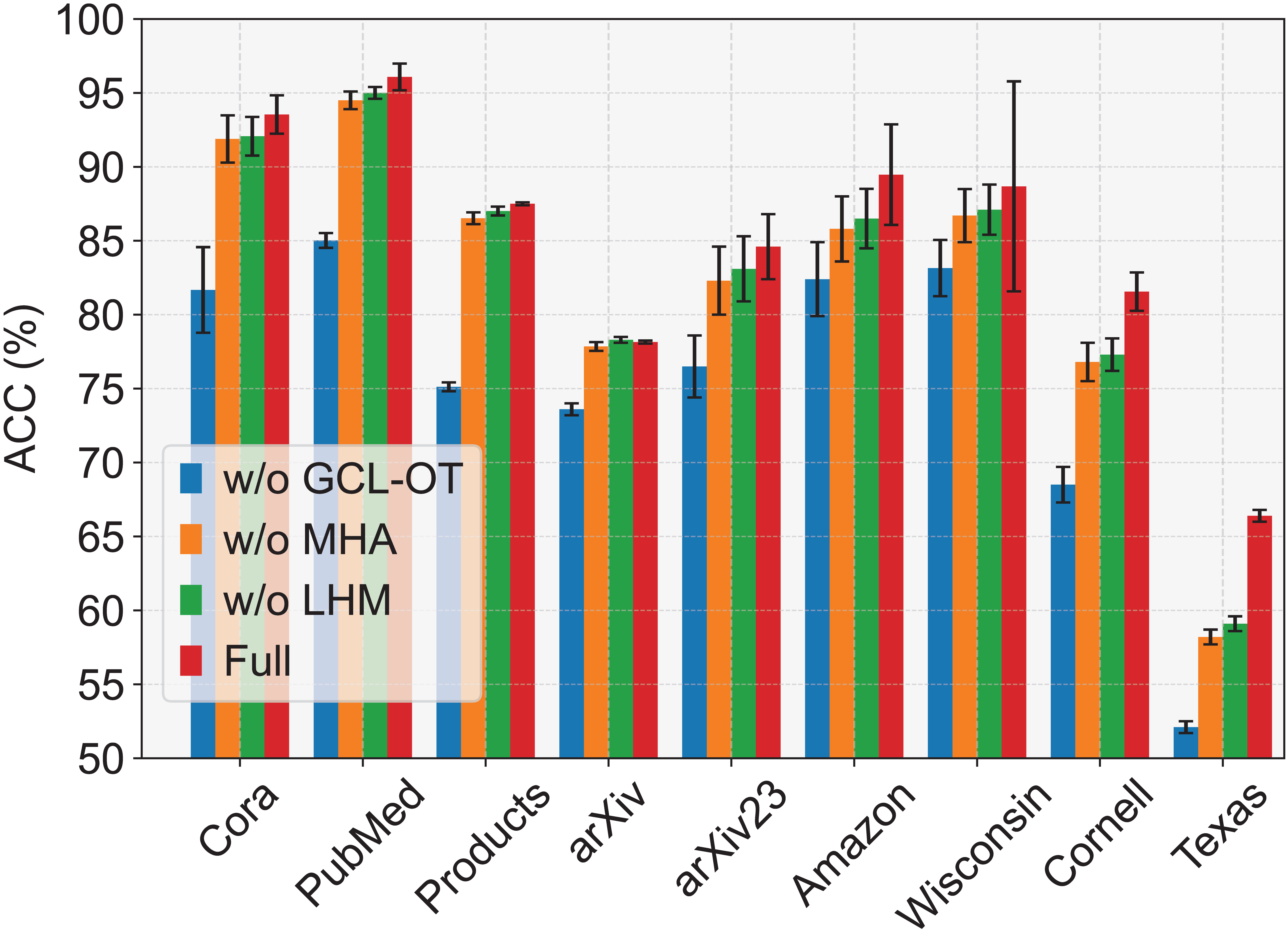}
\caption{Ablation study.}
\label{fig:ablation}
\end{subfigure}
\begin{subfigure}[b]{0.3\linewidth}
\centering
\includegraphics[width=\linewidth]{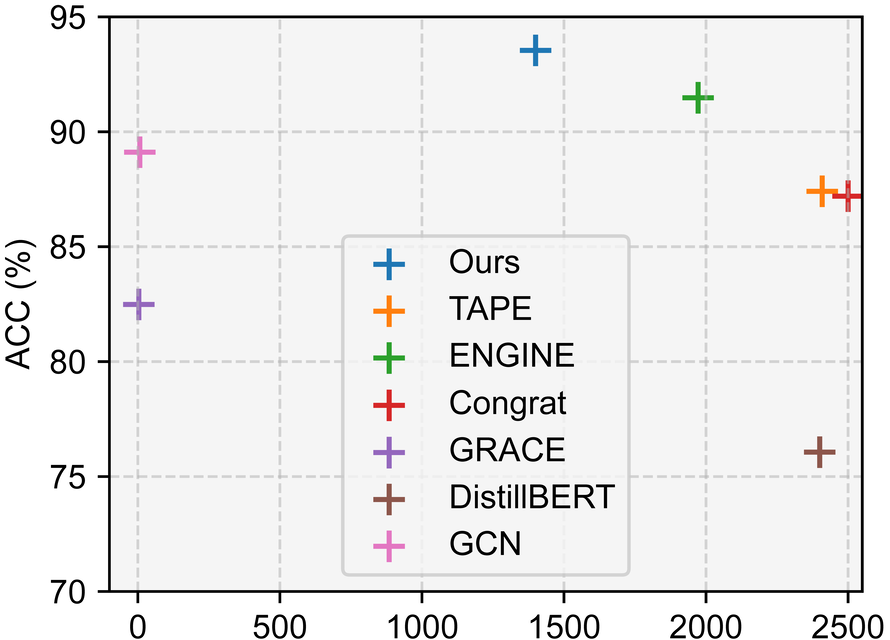}
\caption{Training time (sec.)}
\label{fig:time_acc}
\end{subfigure}
\begin{subfigure}[b]{0.3\linewidth}
\centering
\includegraphics[width=\linewidth]{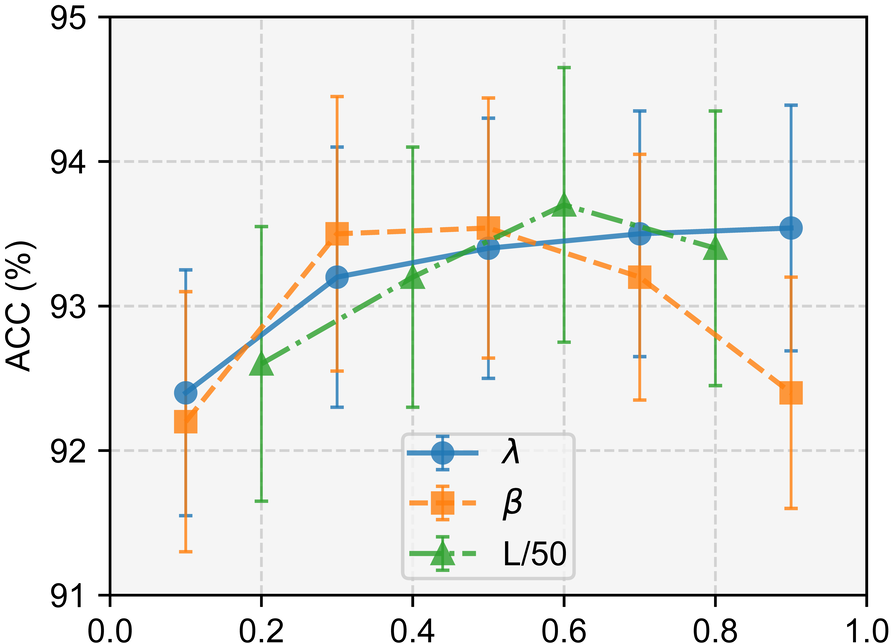}
\caption{Parameters}
\label{fig:parameters}
\end{subfigure}
\caption{Analysis of ablation, computational cost, and hyperparameter sensitivity on Cora.}
\label{fig:hp}
\end{figure*}

\paragraph{Robustness Analysis.}\label{subsubsec:experiments-Robustness}
We evaluate robustness under random edge and text perturbations. Edge perturbation ranges from -500 (removal) to +500 (addition) in increments of 50. Text perturbations range from 0\% to 100\% of nodes, with word perturbation varying from -100\% (removal) to +100\% (addition) at intervals of 10\%. Results on Cora are shown in Figure~\ref{fig:robust}. In the edge perturbation setting, vanilla GCN and GAT suffer sharp accuracy drops under edge deletions, falling approx 50\% and 75\% at the -500 worst condition. At the same time, \ours{} with GCN and \ours{} with GAT achieve relative gains of 24.74\% and 79.44\% respectively. In the text perturbation scenario, DistilBERT shows significant degradation, while our model consistently retains nearly 85\%, exhibiting a smoother performance curve. These experiments highlight the robustness of our model against both structural and text perturbations.

\paragraph{Ablation Study.}\label{subsubsec:experiments-Ablation}
To evaluate the contribution of each component in \ours{}, we conduct an ablation study by selectively contribution three main losses: contrastive loss $\mathcal{L}{\text{GCL-OT}}$, alignment loss $\mathcal{L}_{\text{MHA}}$, and latent homophily mining loss $\mathcal{L}_{\text{LHM}}$. Results in Figure~\ref{fig:ablation} show that removing $\mathcal{L}_{\text{GCL-OT}}$ significantly degrades performance, highlighting its importance for consistency and alignment of structure and text information. Excluding $\mathcal{L}_{\text{MHA}}$ results in significant drops in heterophilic datasets, such as Texas and Cornell. Removing $\mathcal{L}_{\text{LHM}}$ leads to poor performance in datasets like Amazon with weak structural signals. The complete model consistently achieves the best results, confirming the complementary strengths of all components.

\paragraph{Efficiency Comparison.} 
The efficiency of \ours{} is evaluated against baselines by plotting training time against ACC. As shown in Figure~\ref{fig:time_acc}, \ours{} with GCN (blue) achieves the highest ACC while maintaining a relatively short training time, demonstrating a favorable trade-off between efficiency and performance. Although GCN and GRACE train faster, they fail to capture deep textual semantics, resulting in inferior performance. These results highlight the practicality of \ours{} in real-world scenarios.


 \paragraph{PLM Comparison.} 
We further replace the default DistilBERT encoder with stronger PLMs. As shown in~\ref{tab:gclot_plm}, using DeBERTa or RoBERTa yields comparable or slightly better accuracy across most datasets, indicating that \ours{} can benefit from stronger encoders without relying on them.

 \begin{table}[t]
 \centering
\begin{tabular}{@{}lccc@{}}
 \toprule
 & Wisconsin & Cornell & Texas \\
 \midrule
 DeBERTa  & $89.16\pm4.7$ & $87.35\pm6.2$ & $87.30\pm11.8$ \\
 RoBERTa  & $86.42\pm6.6$ & $88.21\pm3.9$ & $86.82\pm14.6$ \\
 \bottomrule
 \end{tabular}
 \caption{\ours{} with different PLMs.}
 \label{tab:gclot_plm}
 \end{table}

\paragraph{Parameter Sensitivity Analysis.} 
\ours{} includes three key hyperparameters: Weight of contrastive loss $\lambda$, temperature of RSM $\beta$, and number of LRSinkhorn iterations. Results with GCN on Cora are shown in Figure~\ref{fig:parameters}. The values of $\lambda$ and $\beta$ are varied within $\{0.0, 0.1, \cdots, 1.0\}$, and the number of LRSinkhorn iterations is tested over $\{10, 20, 30, 40\}$. The results indicate that suitable values of these parameters can enhance performance, and $\lambda$ influences the stability, as reflected in the standard deviation, with only mild fluctuations beyond these values. 

\section{Conclusion}\label{sec:conclusion}

This paper addresses the multi-granular heterophily challenges in TAGs. We propose \ours, a contrastive learning framework with OT with tailored mechanisms to hierarchically align and fuse textual and structural representations. Specifically, for partial heterophily, \ours{} employs the RealSoftMax operator to identify important structure-text relations. For complete heterophily, \ours{} employs a filter prompt to distinguish between alignable and non-alignable embeddings in the transport-based contrastive loss. For latent homophily, \ours{} leverages OT assignments as extra supervision to capture hidden homophilic patterns. Theoretical analysis and extensive experiments in both homophilic and heterophilic settings can demonstrate the effectiveness and robustness of \ours{}.

\section{Acknowledgments}
This paper is supported by the National Key R\&D Program of China (Grant No. 2024YFC3308200).

\bibliography{aaai2026_simple}

\makeatletter
\newif\ifreproStandalone
\reproStandalonetrue
\makeatother

\ifreproStandalone
\setlength{\pdfpagewidth}{8.5in}
\setlength{\pdfpageheight}{11in}
\frenchspacing



\section{Appendix A: Related Work in Detail}\label{app:related_work}

\subsection{OT-based Graph Contrastive Learning}
Optimal transport (OT)~\cite{Monge1781, Villani2009OT}is a mathematical framework for measuring distances between distributions, finding the most cost-efficient way to transform one distribution into another~\cite{lin2023multi, Vincent2022OTGNN, Xu2023GWF}. In Graph Contrastive Learning (GCL), traditional InfoNCE loss usually relies on hard alignment and explicit positive and negative samples, which may lead to representation bias. To this end, researchers begin to use OT to introduce a soft alignment mechanism~\cite{wang2023galopa}. Several methods employ OT to enhance node representation learning, which subsequently benefits downstream tasks such as node clustering ~\cite{Zhang2024CSOT, Wang2024PVCCL,deng2025THESAURUS} and node classification~\cite{xie2024SGEC, Amadou2025FOSSIL, Zhu2022RoSA}.

The node classification method aims to solve the robustness problems caused by non-aligned contrast views and structural differences in the graph by introducing OT distance for soft alignment, subgraph generation, or cross-view comparison, thereby improving the accuracy and stability of node classification~\cite{deng2025THESAURUS,deng2025THESAURUS}. For example, in order to alleviate the uniform effect and cluster assimilation problems, THESAURUS~\cite{deng2025THESAURUS} introduces learnable semantic prototype graphs, employs Gromov–Wasserstein OT to align node embeddings with these prototypes, and uses cross-view prediction of cluster assignment as its learning objective. 
Considering that existing GCL methods often struggle to exploit structural patterns and node similarities fully, FOSSIL~\cite{Amadou2025FOSSIL} introduces an adaptive subgraph generator and formulates a subgraph-level contrastive loss using the Fused Gromov-Wasserstein distance as the similarity measure. 
Beyond clustering and classification, OT has also shown promise in graph few-shot learning~\cite{Liu2025STAR} and node anomaly detection~\cite{Wang2023ACT}.

\subsection{Detailed Baselines}

\begin{itemize}
\item \textbf{MLP}: A multilayer perceptron that uses only the node’s raw text or static embeddings—without graph structure. Serves as a node-classification baseline using only textual semantics.

\item \textbf{DistilBERT}: Directly fine‑tunes a DistilBERT encoder on node text to produce embeddings used for classification. Evaluates performance of strong text-only semantic features if no graph is used.

\item \textbf{GPT‑3.5}: Uses GPT‑3.5 as a zero-shot or few-shot model, or extracts its text embeddings for classification—another text-only baseline.

\item \textbf{GCN}~\cite{kipf2017GCN}: A spectral‑based graph convolutional network that aggregates one‑hop neighborhood features for semi‑supervised node classification.

\item \textbf{GAT}~\cite{velivckovic2018GAT}: Adds multi‑head attention in message passing to adaptively weight neighbor contributions.

\item \textbf{GraphSAGE}~\cite{hamilton2017SAGE}: Uses learned aggregate functions over sampled neighborhoods, supports inductive learning on large graphs.

\item \textbf{GIANT‑BERT}~\cite{chien2022giant}: Learns node textual features by self-supervised neighbor prediction using BERT. These GIANT-generated features are then fed into a GNN classifier.

\item \textbf{GLEM‑DeBERTa}~\cite{zhao2023GLEM}: Trains a language model and GNN iteratively under a variational EM framework, the DeBERTa variant is used to generate text-informed node features that the GNN consumes.

\item \textbf{ENGINE‑LLAMA}~\cite{zhu2024ENGINE}: Introduces an auxiliary, parameterized LM side‑structure to inject graph structure into LLaMA during fine‑tuning, combining graph-aware features with efficient inference and early exit strategies.

\item \textbf{TAPE‑GCN / TAPE‑RevGAT / TAPE‑SAGE}~\cite{he2024TAPE}: First generate node explanations via LLM prompting, those explanations are translated into features that feed into GNN classifiers based on GCN, RevGAT, and GraphSAGE backbones respectively, achieving strong performance via enriched semantics.

\item \textbf{SimTeG‑e5}~\cite{duan2023simteg}: A frustratingly simple two-stage method: fine-tune a text embedding model (e.g., E5) on node classification, and use those embeddings to train a GNN—demonstrating that improved text embeddings alone substantially boost GNN performance.

\item \textbf{LEMP‑TAPE}~\cite{wang2025LEMP4HG}: LEMP‑TAPE extends TAPE by using LEMP4HG’s LM‑enhanced message passing: for selected edges (particularly in heterophilic graph regions), a language model produces a connection analysis given paired node texts, that analysis is encoded and fused via gating with source and target node embeddings to synthesize enriched messages. Only selected edges are enhanced via active learning heuristic (MVRD), reducing overhead while improving robustness on heterophilic graphs.
\item \textbf{DGI}~\cite{velickovic2019dgi}: An unsupervised graph representation learning method that maximizes mutual information between local patch representations and a global graph summary using a GCN encoder. It achieves strong performance on node classification without task-specific labels.
\item \textbf{GRACE}~\cite{zhu2020GRACE}: Learns node embeddings by contrasting augmented views of the graph. GRACE improves notably over DGI, especially on datasets with sparse node features, by exploiting topology-aware negative sampling and graph augmentations.
\item \textbf{ConGraT}~\cite{brannon2024congrat}: A self‑supervised method jointly pretraining a language model and a GNN via contrastive alignment of text and node representations in the same latent space, balancing textual and structural signals for downstream tasks.
\end{itemize}

\section{Appendix B: Proof}\label{app:proof}

\subsection{Proof of Proposition 1}\label{app:Proposition1}

\begin{proof}
For a batch of size $N$, given the similarity $\bar{s}{ij}$ and temperature $\tau>0$, OT assignment $\sum_j q^*_{ij} = 1$, $\sum_i q^*_{ij} = 1$, $0 \leq q^*_{ij} \leq 1$, 
{\small
\begin{equation}
\begin{aligned}
&\mathcal{L}_{\text{MHA}}=-\mathbb{E}_{i}\left[\log\frac{{d_{ii}}}{\sum_{j=1}^{r}d_{ij}}+\log\frac{d_{ii}}{\sum_{j=1}^r d_{ji}}\right]
\\
&= -\mathbb{E}_{i}\left[\log\frac{\exp(q^*_{ii} \bar{s}_{ii}/\tau)}{\sum_{j=1}^r\exp( q^*_{ij} \bar{s}_{ij}/\tau)}+\log\frac{\exp(q^*_{ii}\bar{s}_{ii} /\tau)}{\sum_{j=1}^r\exp( q^*_{ji} \bar{s}_{ji} /\tau)}\right]
\\
&= -\mathbb{E}_{i}\left[\log\frac{\exp(q^*_{ii} s_{ii}/\tau)}{\sum_{j=1}^r\exp(q^*_{ij} s_{ij}/\tau)}+\log\frac{\exp(q^*_{ii} s_{ii} /\tau)}{\sum_{j=1}^r\exp(q^*_{ji} s_{ji} /\tau)}\right]
\\
&\leq -\mathbb{E}_{i}\left[\log\frac{\exp(q^*_{ii} s_{ii}/\tau)}{\sum_{j=1}^N\exp( q^*_{ij} s_{ij}/\tau)}+\log\frac{\exp(q^*_{ii} s_{ii} /\tau)}{\sum_{j=1}^N\exp( q^*_{ji} s_{ji} /\tau)}\right]
\\
&\leq -\mathbb{E}_{i}\left[\log\frac{\exp(s_{ii}/\tau)}{\sum_{j=1}^N\exp(s_{ij}/\tau)}+\log\frac{\exp(s_{ii} /\tau)}{\sum_{j=1}^N\exp(s_{ji} /\tau)}\right]
\\
&=\mathcal{L}_{\text{InfoNCE}}.
\end{aligned}
\end{equation}
}
Thus we obtain that the upper bound for $\mathcal{L}_{\text{MHA}}$ has the form of $-\mathbb{E}_{i}\left[\log\frac{\exp(s_{ii}/\tau)}{\sum_{j=1}^N\exp(s_{ij}/\tau)}+\log\frac{\exp(s_{ii}/\tau)}{\sum_{j=1}^N\exp(s_{ji}/\tau)}\right]$, which is the same as the objective in the CLIP-style symmetric InfoNCE, i.e., the sum of the two directions~\cite{oord2018representation, poole2019variational, radford2021learning,zhu2020GRACE,zhu2025GraphCLIP}. 

The variational mutual information (MI) in the InfoNCE view is
\begin{equation}
\operatorname{MI}(\mathbf{H}^{\mathcal{N}},\mathbf{H}^{\varpi}) \geq \log N - \mathcal{L}_{\text{InfoNCE}},
\end{equation}
where $N$ is the number of negative sample. When we want to minimize $\mathcal{L}_{\text{MHA}}$ for optimization, it is quite effective to minimize $\mathcal{L}_{\text{InfoNCE}}$, actually maximizing $\log N - \mathcal{L}_{\text{InfoNCE}}$, thereby indirectly maximizing the lower bound of the mutual information $\operatorname{MI}(\mathbf{H}^{\mathcal{N}},\mathbf{H}^{varpi})$. 
\end{proof}

\subsection{Proof of Proposition 2}\label{app:proof_LHM}
\begin{proof}
\begin{equation}
\begin{aligned}
&\mathcal{L}_{\text{LHM}}\\
&=-\frac{1}{N} \sum_{i=1}^N \sum_{j=1}^r p_{ij}(\log \hat{k}_{ij}^{'} + \log \hat{k}^{''}_{ij})\\
&=-\mathbb{E}_{ij} \left[ p_{ij}(\log \hat{k}_{ij}^{'} + \log \hat{k}^{''}_{ij})\right]
\\
&=-\mathbb{E}_{ij} \left[p_{ij}(\log(\mathrm{softmax}_j(\hat{s}_{ij}/\tau)) + \log(\mathrm{softmax}_i(\hat{s}_{ij}/\tau))\right]
\\
&=-\mathbb{E}_{ij} [ p_{ij}g(\log\frac{\exp(\hat{s}_{ij}/\tau}{\sum_{k}\exp(\hat{s}_{ik}/\tau)} \\&\quad+ \log\frac{\exp(\hat{s}_{ij}/\tau)}{\sum_{k}\exp(\hat{s}_{kj}/\tau)}g)]
\\
&=-\mathbb{E}_{ij} g[(\delta_{ij} + q^{*}_{ij})(\log\frac{\exp(\hat{s}_{ij}/\tau}{\sum_{k}\exp(\hat{s}_{ik}/\tau)} \\&\quad + \log\frac{\exp(\hat{s}_{ij}/\tau)}{\sum_{k}\exp(\hat{s}_{kj}/\tau)})g]
\\
&=-\mathbb{E}_{ij} g[(\delta_{ij} + q^{*}_{ij})(\log\frac{\exp(\hat{s}_{ij}/\tau}{\sum_{k}\exp(\hat{s}_{ik}/\tau)} \\&\quad+ \log\frac{\exp(\hat{s}_{ij}/\tau)}{\sum_{k}\exp(\hat{s}_{kj}/\tau)})g]\\
&\leq-\mathbb{E}_{ij} \left[\log\frac{\exp(\hat{s}_{ii}/\tau)}{\sum_{k}\exp(\hat{s}_{ik}/\tau)}+\log\frac{\exp(\hat{s}_{ii} /\tau)}{\sum_{k}\exp(\hat{s}_{ki} /\tau)}\right]
\\
\\
&= \mathcal{L}_{\text{InfoNCE}}
\end{aligned}
\end{equation}

Assume diagonal dominance along both the i-th row and column, we have $\hat{s}_{ii}=\max_j \hat{s}_{ij},\hat{s}_{ii}=\max_i \hat{s}_{ij}, \forall i \in [0,1]$. Thus we obtain that the upper bound for $\mathcal{L}_{\text{LHM}}$ has the form of $-\mathbb{E}_{i}\left[\log\frac{\exp(\hat{s}_{ii}/\tau)}{\sum_{j=1}^N\exp(\hat{s}_{ij}/\tau)}+\log\frac{\exp(\hat{s}_{ii}/\tau)}{\sum_{j=1}^N\exp(\hat{s}_{ji}/\tau)}\right]$, which is also the same as the objective in the CLIP-style symmetric InfoNCE, and the variational MI in InfoNCE is
\begin{equation}
\operatorname{MI}(\mathbf{H}^{\zeta},\mathbf{H}^{t}) \geq \log N - \mathcal{L}_{\text{InfoNCE}},
\end{equation}
where $N$ is the number of negative sample. When we want to minimize $\mathcal{L}_{\text{LHM}}$ for optimization, it is quite effective to minimize $\mathcal{L}_{\text{InfoNCE}}$, actually maximizing $\log N - \mathcal{L}_{\text{InfoNCE}}$, thereby indirectly maximizing the lower bound of the mutual information $\operatorname{MI}(\mathbf{H}^{\zeta},\mathbf{H}^{t})$. 
\end{proof}

\subsection{Proof of Theorem 1}
\begin{proof}
Let $\mathbf{z}_i = \alpha_1 \mathbf{h}_i^{\zeta} + \alpha_2 \mathbf{h}_i^{t}$, and define the final embedding as $\mathbf{h}_i = \mathbf{W} \mathbf{z}_i$, where $\mathbf{W}$ is a linear transformation with full column rank. Then the map $\mathbf{z}_i \mapsto \mathbf{h}_i$ is injective, and admits a left inverse on the range of $\mathbf{W}$. Since mutual information is invariant to bijective transformations, we have:
\begin{equation}
\begin{aligned}
MI(\mathbf{x}_i^{(k)}, \mathbf{h}_i) = MI(\mathbf{x}_i^{(k)}, \mathbf{z}_i).
\end{aligned}
\end{equation}
By the data-processing inequality, no post-processing can increase information, and injective maps preserve it. Therefore, the information preserved in $(\mathbf{h}_i^{\zeta}, \mathbf{h}_i^{t})$ is maintained in $\mathbf{h}_i$.

Minimizing the joint loss $\mathcal{L}_{\text{GCL-OT}} = \mathcal{L}_{\text{MHA}} + \mathcal{L}_{\text{LHM}}$ maximizes variational lower bounds of $MI(\mathbf{H}^{\zeta}, \mathbf{H}^{t}))$, and by injectivity of the linear head, the resulting final representation $\mathbf{h}_i$ preserves mutual information with the input neighborhood $\mathbf{x}_i^{(k)}$, i.e., $MI(\mathbf{x}_i^{(k)}, \mathbf{h}_i)$ increases as well.
\end{proof}

\subsection{Proof of Remark} 

For two distributions $P$ and $Q$ with mixture $M = \tfrac{1}{2}(P+Q)$, the Jensen-Shannon divergence is
\begin{equation}
\begin{aligned}
\operatorname{JS}(P,Q)=\frac{1}{2}\operatorname{KL}(P\|M) + \frac{1}{2}\operatorname{KL}(Q\|M).
\end{aligned}
\end{equation}

Consider a binary classifier $D:\mathcal{Z}\to(0,1)$ that distinguishes samples from $P$ or $Q$ with equal priors. Its cross-entropy loss is
\begin{equation}
\begin{aligned}
\mathcal{L}(D) = \mathbb{E}_{z\sim P} [-\log D(z)]+\mathbb{E}_{z\sim Q} [-\log (1-D(z)) ].
\end{aligned}
\end{equation}
It is well known that the optimal discriminator is
\begin{equation}
\begin{aligned}
D^{*}(z)=\frac{P(z)}{P(z)+Q(z)},
\end{aligned}
\end{equation}
and the optimal value satisfies
\begin{equation}
\mathcal{L}(D^{*}) = 2\log 2 - 2 \operatorname{JS}(P,Q).
\end{equation}
Equivalently, for any $D$,
\begin{equation}
\mathcal{L}(D) \ge 2\log 2 - 2 \operatorname{JS}(P,Q),
\end{equation}
with equality at $D^{*}$.

Fix a view $i\in\{\zeta,t\}$. Under the standard one-positive–many-negatives construction with $r$ candidates, we draw
\begin{equation}
\begin{aligned}
Z\mid Y=\text{pos} \sim P_{\text{pos}}^{i}, 
Z\mid Y=\text{neg} \sim P_{\text{neg}}^{i},
\end{aligned}
\end{equation}
with priors
\begin{equation}
\begin{aligned}
\pi_{\text{pos}}=\frac{1}{r},
\pi_{\text{neg}}=\frac{r-1}{r}.
\end{aligned}
\end{equation}

Let the OT-weighted similarity critic be
\begin{equation}
\begin{aligned}
g^{i}(u,v) = \frac{q^{*}_{uv} s_{uv}}{\tau},
\end{aligned}
\end{equation}
and define the induced discriminator
\begin{equation}
\begin{aligned}
D^{i}(Z) = \sigma (g^{i}(Z)),
\end{aligned}
\end{equation}
where $\sigma(\cdot)$ is the logistic sigmoid. As shown in prior work on multi-view, the InfoNCE-type loss can be written as a binary cross-entropy between positive and negative pairs plus a prior-dependent constant. 
\begin{equation}
\begin{aligned}
\mathcal{L}^{i} = \mathbb{E}_{Z\sim P_{\text{pos}}^{i}} [-\log D^{i}(Z)]\\
+\mathbb{E}_{Z\sim P_{\text{neg}}^{i}} [-\log (1-D^{i}(Z))]+\log r.
\end{aligned}
\end{equation}

Minimizing over $D^{i}$ yields the same optimal form but with prior $1/r$, which only changes the constant. Thus there exists an optimal discriminator $D^{i,*}$ such that
\begin{equation}
\begin{aligned}
\mathcal{L}^{i} \ge 2\log r - 2 \operatorname{JS}^{i},\\
\operatorname{JS}^{i}=\operatorname{JS} (P_{\text{pos}}^{i},P_{\text{neg}}^{i} ),
\end{aligned}
\end{equation}
with equality at $D^{i,*}$. 
This derivation does not depend on the specific critic parameterization, such as dot-product or weighted addition, only on the expressiveness of $g^{i}$. Hence, it applies directly to GCL-OT.

The full GCL-OT loss sums bi-directional terms from both views:
\begin{equation}
\begin{aligned}
\mathcal{L}_{\text{GCL-OT}} = \mathcal{L}^{\zeta}+ \mathcal{L}^{t}.
\end{aligned}
\end{equation}
Each bi-directional term consists of symmetric softmaxes in row-wise and column-wise,
\begin{equation}
\begin{aligned}
\mathcal{L}^{i}_{\text{GCL-OT}}=\mathcal{L}^{i,\text{row}}+\mathcal{L}^{i,\text{col}}, i\in\{\zeta,t\}.
\end{aligned}
\end{equation}
By the argument applied to each direction,
\begin{equation}
\begin{aligned}
\mathcal{L}^{i,\text{row}}_{\text{GCL-OT}} \ge 2\log r - 2 \operatorname{JS}^{i}, \mathcal{L}^{i,\text{col}}_{\text{GCL-OT}} \ge 2\log r - 2 \operatorname{JS}^{i}.
\end{aligned}
\end{equation}
Summing the two directions and collecting constants yields
\begin{equation}
\begin{aligned}
\mathcal{L}_{\text{GCL-OT}} \ge 4\log r - 2 (\operatorname{JS}^{\zeta}+\operatorname{JS}^{t}).
\end{aligned}
\end{equation}
Therefore, under optimal OT-based discriminators, minimizing $\mathcal{L}_{\text{GCL-OT}}$ increases the JSD between positive and negative pair distributions in both views, in addition to maximizing the underlying mutual information.

\subsection{Proof of Connection with Downstream Tasks}\label{app:downstream}
\begin{proof}
The Bayes error of the optimal classifier denotes as $P_e(\mathbf{H})$
\begin{equation}
 P_e(\mathbf{H}) = \mathbb{E}_{\mathbf{H}}[1 - \max_{c \in \mathcal{Y}} p(y=c \mid \mathbf{H})].
\end{equation}
We also consider the Bayes errors $P_e(\mathbf{H}^{\zeta})$ and $P_e(\mathbf{H}^{t})$ of representations learned from the structural view and textual view:
\begin{equation}
 \sup P_e(\mathbf{H}) \triangleq H(y \mid \mathbf{H}),
\end{equation}
and analogously for $\mathbf{H}^{\zeta}$ and $\mathbf{H}^{t}$.

\begin{lemma}[Bayes error and conditional entropy]
Assume $0 \le P_e(\mathbf{H}) \le 1 - 1/|\mathcal{Y}|$.
Then
\begin{equation}
 -\log (1 - P_e(\mathbf{H}))\le H(y \mid \mathbf{H}).
\end{equation}
In particular, $\sup P_e(\mathbf{H}) = H(y \mid \mathbf{H})$ is an upper bound on the Bayes error $P_e(\mathbf{H})$.
\end{lemma}
This is a standard consequence of Fano-type inequalities that relate classification error to conditional entropy.

By the chain rule of entropy, we also have the identity $H(y \mid \mathbf{H}) = H(y) - I(\mathbf{H}, y),$ which shows that reducing the Bayes-error upper bound is equivalent to increasing the mutual information between the representation $\mathbf{H}$ and the downstream label $y$.

Let $ z^{\zeta} = g^{\zeta}_{\theta}(\mathcal{G}), z^{t} = g^{t}_{\theta}(\mathbf{X})$ denote the corresponding view-specific self-supervised signals, which induce the positive/negative pair distributions $P_{\text{pos}}^{\zeta}, P_{\text{neg}}^{\zeta}$ and $P_{\text{pos}}^{t}, P_{\text{neg}}^{t}$ used in $\mathcal{L}_{\text{GCL-OT}}$. The baseline losses $\mathcal{L}_{\text{InfoNCE}}$ and $\mathcal{L}_{\text{NC}}$ operate on $z^{\zeta}$ and $z^{t}$ separately, yielding representations $\mathbf{H}^{\zeta}$ and $\mathbf{H}^{t}$, while $\mathcal{L}_{\text{GCL-OT}}$ uses the joint signal $(z^{\zeta}, z^{t})$ to learn an aligned representation $\mathbf{H}$. Let $\mathbf{H}$ be the representation learned by minimizing $\mathcal{L}_{\text{GCL-OT}}$, and let $\mathbf{H}^{\zeta}$ and $\mathbf{H}^{t}$ be the representations learned by minimizing $\mathcal{L}_{\text{InfoNCE}}$ and $\mathcal{L}_{\text{NC}}$ alone, respectively. Then
\begin{equation}
 I(\mathbf{H}, y) \ge I(\mathbf{H}^{\zeta}, y), I(\mathbf{H}, y)\ge I(\mathbf{H}^{t}, y).
\end{equation}

Let $\mathbf{H}^{\text{sup}}$ denote an ideal supervised representation trained with access to labels $y$. Decomposing the information carried by the self-supervised signals yields
\begin{equation}
 I(\mathbf{H}, y) = I(\mathbf{H}^{\text{sup}}, y) - I ((\mathcal{G},\mathbf{X}), y | z^{\zeta}, z^{t} ).
\end{equation}
The second term on the right-hand side measures the information gap between the ideal supervised signal and the joint self-supervised signals, by nonnegativity of conditional mutual information,
\begin{equation}
I ((\mathcal{G},\mathbf{X}), y | z^{\zeta}, z^{t} ) \ge 0.
\end{equation}
If we only use the structural view or only the textual view, the corresponding decompositions are
\begin{equation}
 I(\mathbf{H}^{\zeta}, y) =I(\mathbf{H}^{\text{sup}}, y) - I ((\mathcal{G},\mathbf{X}), y | z^{\zeta} ),
 \\
 I(\mathbf{H}^{t}, y)=I(\mathbf{H}^{\text{sup}}, y) - I ((\mathcal{G},\mathbf{X}), y | z^{t} ).
\end{equation}
Using the fact that conditioning reduces conditional mutual information, we can obtain
\begin{equation}
 I ((\mathcal{G},\mathbf{X}), y | z^{\zeta} )\ge I ((\mathcal{G},\mathbf{X}), y | z^{\zeta}, z^{t} ),
\end{equation}
and similarly for $z^{t}$. Plugging these into the expressions above yields
\begin{equation}
 I(\mathbf{H}, y) \ge I(\mathbf{H}^{\zeta}, y), I(\mathbf{H}, y) \ge I(\mathbf{H}^{t}, y).
\end{equation}

Based on the above results, under the same assumptions as above, we can obtain a tighter Bayesian error bound:
\begin{equation}
 H(y \mid \mathbf{H}) \le H(y \mid \mathbf{H}^{\zeta}),
 H(y \mid \mathbf{H}) \le  H(y \mid \mathbf{H}^{t}).
\end{equation}
Equivalently, in terms of the Bayes-error upper bounds,
\begin{equation}
 \sup P_e(\mathbf{H}) \le  \min (\sup P_e(\mathbf{H}^{\zeta}),\sup P_e(\mathbf{H}^{t}) ).
\end{equation}
Hence, the representation learned by minimizing $\mathcal{L}_{\text{GCL-OT}}$ admits a no-looser (and typically tighter) upper bound on the downstream Bayes error than representations learned by optimizing $\mathcal{L}_{\text{InfoNCE}}$ or $\mathcal{L}_{\text{NC}}$ alone, or their simple combination without OT-based alignment.Consequently, downstream node classification tasks can benefit from the representations learned from \ours{}.
\end{proof}

\subsection{Proof of RealSoftMax}\label{app:RealSoftMax}

Given 
\begin{equation}
\begin{aligned}
s_{ij} = \frac{1}{2}(\mathbb{E}_{k=1}^{K}[\operatorname{RSM}_{\beta}(\{\mathbf{h}^{\mathcal{N}}_{ik} \cdot \mathbf{h}^{\varpi}_{jw}\}_{w=1}^{W})]\\+\mathbb{E}_{w=1}^{W}[\operatorname{RSM}_{\beta} (\{\mathbf{h}^{\varpi}_{iw} \cdot \mathbf{h}^{\mathcal{N}}_{jk}\}_{k=1}^{K})]),
\end{aligned}
\label{eq:realSoftMax}
\end{equation}
where $\mathbf{h}^{\mathcal{N}}_{ik} \in \mathbb{R}^d$ is the $k$-th neighbor structural embedding of node $i$, $\mathbf{h}^{\varpi}_{iw} \in \mathbb{R}^d$ is the $w$-th textual token text embedding of node $i$, $\operatorname{RSM}_{\beta}(\cdot)$ is the RealSoftMax operator with temperature $\beta$, $\mathbb{E}_{k=1}^{K}[\cdot] = \frac{1}{K} \sum_{k=1}^{K}(\cdot)$ denotes uniform averaging over $k$. 

\begin{proof}
Let $\{x_w\}_{w=1}^{W} = \{\mathbf{h}^{\mathcal{N}}_{ik} \cdot \mathbf{h}^{\varpi}_{jw}\}_{w=1}^{W})$, and $\{x_k\}_{k=1}^{K} =\{\mathbf{h}^{\varpi}_{iw}\cdot \mathbf{h}^{\mathcal{N}}_{jk}\}_{k=1}^{K}$, $x^*_w$ and $x^*_k$ are the max value of $\{x_w\}_{w=1}^{W}$ and $\{x_k\}_{k=1}^{K}$, respectively.

\paragraph{Case 1: $\beta \to 0^+$.} In this regime, the exponential function sharply amplifies the largest $x_l$, and we have
\begin{equation}
\begin{aligned}
\lim_{\beta \to 0^{+}} \sum_{w=1}^N \exp( \frac{x_w}{\beta} ) = \exp( \frac{\max_w x_w}{\beta} )
\end{aligned}
\end{equation}
leading to
\begin{equation}
\lim_{\beta \to 0^{+}} \operatorname{RSM}_\beta(\{x_w\}) = x^*_w.
\end{equation}

similarity, we have
\begin{equation}
\lim_{\beta \to 0^{+}} \operatorname{RSM}_\beta(\{x_k\}) = x^*_k.
\end{equation}

\paragraph{Case 2: $ \beta \to +\infty$.} 
For each $l$, the first-order Taylor approximation of the exponential function at $0$:
\begin{equation}\begin{aligned}
\exp (\frac{x_l}{\beta}) = 1+\frac{x_l}{\beta}+o (\frac{1}{\beta}), \quad \text{as} \quad \beta \to \infty,
\end{aligned}\end{equation}
where $o (\frac{1}{\beta})$ denotes higher-order infinitesimals. Summing over $l$ gives
\begin{equation}\begin{aligned}
\sum_{l=1}^N \exp (\frac{x_l}{\beta}) = N+\frac{1}{\beta} \sum_{l=1}^N x_l+o (\frac{1}{\beta}).
\end{aligned}\end{equation}

Applying the Taylor expansion of $\log(1+z)$ around $z=0$, $ \log(1+z) = z - \frac{z^2}{2}+o(z^2),$ 
we obtain
\begin{equation}
\begin{aligned}
\log (N+\frac{1}{\beta} \sum_{l=1}^N x_l)
=\log N+\log (1+\frac{1}{N\beta} \sum_{l=1}^N x_l)\\
=\log N+\frac{1}{N\beta} \sum_{l=1}^N x_l+o (\frac{1}{\beta}).
\end{aligned}
\end{equation}

Multiplying both sides by $\beta$, we get
\begin{equation}
\begin{aligned}
\beta \log (\sum_{l=1}^N \exp (\frac{x_l}{\beta}))= \beta \log N+\frac{1}{N} \sum_{l=1}^N x_l+o(1),
\end{aligned}
\end{equation}
where $o(1)$ denotes a term vanishing as $\beta \to \infty$.

Thus, in the limit $\beta \to \infty$, we obtain
\begin{equation}
\begin{aligned}
\lim_{\beta \to \infty} (\beta\log(\sum_{l=1}^N \exp (\frac{x_l}{\beta})) - \beta \log N)= \frac{1}{N} \sum_{l=1}^N x_l.
\end{aligned}
\end{equation}

Thus, we have
\begin{equation}
\begin{aligned}
\beta \log( \sum_{w=1}^{W} \exp( \frac{\mathbf{h}^{\mathcal{N}}_{ik} \cdot \mathbf{h}^{\varpi}_{jw}}{\beta}))
\sim \\
\begin{cases}
\max\limits_w ( \mathbf{h}^{\mathcal{N}}_{ik} \cdot \mathbf{h}^{\varpi}_{jw} ), & \beta \to 0 \\
\beta \log W + \frac{1}{W} \sum\limits_{w=1}^{W} ( \mathbf{h}^{\mathcal{N}}_{ik} \cdot \mathbf{h}^{\varpi}_{jw} ), & \beta \to \infty
\end{cases}
\end{aligned}
\end{equation}
Similarly, $\forall w \in \{1, \dots, W\}$, the term $\beta\log(\sum_{k=1}^{K}\exp(\frac{\mathbf{h}^{\varpi}_{iw} \cdot \mathbf{h}^{\mathcal{N}}_{jk}}{\beta}))$ satisfies
\begin{equation}
\begin{aligned}
\beta\log(\sum_{k=1}^{K}\exp(\frac{\mathbf{h}^{\varpi}_{iw} \cdot \mathbf{h}^{\mathcal{N}}_{jk}}{\beta}))\sim\\
\begin{cases}
\max\limits_k ( \mathbf{h}^{\varpi}_{iw} \cdot \mathbf{h}^{\mathcal{N}}_{jk}), & \beta \to 0 \\
\beta \log K + \frac{1}{K} \sum\limits_{k=1}^{K}( \mathbf{h}^{\varpi}_{iw} \cdot \mathbf{h}^{\mathcal{N}}_{jk}), & \beta \to \infty
\end{cases}
\end{aligned}
\end{equation}

Therefore, as $\beta\to0$, we have
\begin{equation}
\begin{aligned}
\lim_{\beta \to 0} s_{ij} = \frac{1}{2} (\frac{1}{K} \sum_{k=1}^{K} \max_{w} (\mathbf{h}^{\mathcal{N}}_{ik} \cdot \mathbf{h}^{\varpi}_{jw}) 
\\
+ \frac{1}{W} \sum_{w=1}^{W} \max_{k} (\mathbf{h}^{\varpi}_{iw} \cdot \mathbf{h}^{\mathcal{N}}_{jk})),
\end{aligned}
\end{equation}
which averages the maximal token-neighbor similarity for each neighbor and each token.

Meanwhile, as $\beta \to \infty$, ignoring the additive constants $\beta \log K$ and $\beta \log W$, we obtain
\begin{equation}
\begin{aligned}
\lim_{\beta \to \infty}(s_{ij} - \frac{\beta}{2}(\log K+\log W))\\
= \frac{1}{KW} \sum_{k=1}^{K} \sum_{w=1}^{W} (\mathbf{h}^{\mathcal{N}}_{ik} \cdot \mathbf{h}^{\varpi}_{jw}),
\end{aligned}
\end{equation}
which is simply the mean of all neighbor-token similarities.

This completes the proof.
\end{proof}

\subsection{Derivation of the OT Distance Solved by LRSinkhorn Algorithm}\label{app:ot}
The Low-Rank Sinkhorn (LRSinkhorn) algorithm is a scalable variant of the Sinkhorn algorithm for entropic optimal transport (OT). It preserves the alternating scaling updates of Sinkhorn while replacing the full Gibbs kernel with a low-rank approximation to reduce complexity.

We begin with the entropic regularized OT problem:
\begin{equation}
W_\varepsilon(\mu, \nu) = \min_{\pi \in \Pi(\mu, \nu)} \langle C, \pi \rangle + \varepsilon \operatorname{KL}(\pi \Vert \mu \otimes \nu),
\end{equation}
where $ \Pi(\mu, \nu) $ denotes the set of couplings with marginals $ \mu \in \Delta^n $ and $ \nu \in \Delta^m $, and $ C \in \mathbb{R}^{n \times m} $ is the cost matrix.

The solution $ \pi^\star $ to this problem has the factorized form:
\begin{equation}
\pi^\star = \mathrm{diag}(a) K \mathrm{diag}(b), \quad \text{with} \quad K = \exp(-C / \varepsilon),
\end{equation}
and the vectors $ a \in \mathbb{R}^n $, $ b \in \mathbb{R}^m $ are obtained by the Sinkhorn iterations:
\begin{align}
b &\leftarrow \nu \oslash (K^\top a), \\
a &\leftarrow \mu \oslash (K b),
\end{align}
where $ \oslash $ denotes element-wise division.

\paragraph{Low-Rank Kernel Approximation.}
For moderate values of $ \varepsilon $, the Gibbs kernel $ K $ is often numerically low-rank. One may approximate it as:
\begin{equation}
K \approx U V^\top, \quad U \in \mathbb{R}^{n \times r}, V \in \mathbb{R}^{m \times r}, r \ll \min(n, m),
\end{equation}
using Nyström approximation, randomized SVD, or similar techniques.

\paragraph{Low-Rank Sinkhorn Iterations.}
Using this low-rank decomposition, the Sinkhorn iterations become:

\begin{itemize}
\item Compute $ s = V^\top b \in \mathbb{R}^r $ and update $ a = \mu \oslash (U s) $.
\item Compute $ t = U^\top a \in \mathbb{R}^r $ and update $ b = \nu \oslash (V t) $.
\end{itemize}

Each iteration now costs $ \mathcal{O}((n + m)r) $, significantly reducing the original $ \mathcal{O}(nm) $ complexity.

The iterations still solve the same fixed-point equations. The only approximation lies in the low-rank factorization of $ K $. Provided $ r $ is chosen proportional to the effective rank of $ K $, the resulting plan $ \hat{\pi} $ satisfies $\langle C, \hat{\pi} \rangle \leq W_\varepsilon(\mu, \nu) + \delta$, for any desired precision $ \delta > 0 $. 

LRSinkhorn enables efficient computation of entropic OT on large-scale problems by leveraging the low-rank structure of the Gibbs kernel.

\begin{table*}[htbp]
\centering
\small
\begin{tabular}{@{}lp{14cm}@{}}
\toprule
Dataset & Prompt \\
\midrule
Cora & Abstract: $\langle$ abstract text $\rangle$ $\backslash$n Title: $\langle$ title text $\rangle$ $\backslash$n Question: Which of the following sub-categories of AI does this paper belong to: Case Based, Genetic Algorithms, Neural Networks, Probabilistic Methods, Reinforcement Learning, Rule Learning, Theory? If multiple options apply, provide a comma-separated list ordered from most to least related, then for each choice you gave, explain how it is present in the text. $\backslash$n $\backslash$n Answer: \\
\midrule
Pubmed & Abstract: $\langle$ abstract text $\rangle$ $\backslash$n Title: $\langle$ title text $\rangle$ $\backslash$n Question: Does the paper involve any cases of Type 1 diabetes, Type 2 diabetes, or Experimentally induced diabetes? Please give one or more answers of either Type 1 diabetes, Type 2 diabetes, or Experimentally induced diabetes, if multiple options apply, provide a comma-separated list ordered from most to least related, then for each choice you gave, give a detailed explanation with quotes from the text explaining why it is related to the chosen option. $\backslash$n $\backslash$n Answer: \\
\midrule
arXiv & Abstract: $\langle$ abstract text $\rangle$ $\backslash$n Title: $\langle$ title text 
$\rangle$ $\backslash$n Question: Which arXiv CS subcategory does this paper belong to? Give 5 
likely arXiv CS sub-categories as a comma-separated list ordered from most to least likely, in the 
form “cs.XX”, and provide your reasoning. $\backslash$n $\backslash$ n Answer: \\
\midrule
Products & Product description: $\langle$ product description $\rangle$ $\backslash$n Question: 
Which of the following category does this product belong to: 1) Home \& Kitchen, 2) Health \& 
Personal Care, 3) Beauty, 4) Sports \& Outdoors, 5) Books, 6) Patio, Lawn \& Garden, 7) Toys \& 
Games, 8) CDs \& Vinyl, 9) Cell Phones \& Accessories, 10) Grocery \& Gourmet Food, 11) Arts, 
Crafts \& Sewing, 12) Clothing, Shoes \& Jewelry, 13) Electronics, 14) Movies \& TV, 15) Software, 
16) Video Games, 17) Automotive, 18) Pet Supplies, 19) Office Products, 20) Industrial \& 
Scientific, 21) Musical Instruments, 22) Tools \& Home Improvement, 23) Magazine Subscriptions, 
24) Baby Products, 25) NAN, 26) Appliances, 27) Kitchen \& Dining, 28) Collectibles \& Fine Art, 
29) All Beauty, 30) Luxury Beauty, 31) Amazon Fashion, 32) Computers, 33) All Electronics, 34) 
Purchase Circles, 35) MP3 Players \& Accessories, 36) Gift Cards, 37) Office \& School Supplies, 
38) Home Improvement, 39) Camera \& Photo, 40) GPS \& Navigation, 41) Digital Music, 42) Car 
Electronics, 43) Baby, 44) Kindle Store, 45) Kindle Apps, 46) Furniture \& Decor? Give 5 likely 
categories as a comma-separated list ordered from most to least likely, and provide your 
reasoning. $\backslash$n $\backslash$n Answer: \\
\midrule
arXiv23 & Abstract: $\langle$ abstract text $\rangle$ $\backslash$n Title: $\langle$ title text 
$\rangle$ $\backslash$n Question: Which arXiv CS subcategory does this paper belong to? Give 5 
likely arXiv CS sub-categories as a comma-separated list ordered from most to least likely, in the 
form “cs.XX”, and provide your reasoning. $\backslash$n $\backslash$n Answer: \\
\midrule
Cornell & Webpage description: $\langle$ webpage description $\rangle$ $\backslash$n Question: Which of the following categories does this webpage belong to: 1) department, 2) project, 3) faculty, 4) course, 5) student, 6) other? If multiple options apply, provide a comma-separated list ordered from most to least related, then for each choice you gave, provide your reasoning. $\backslash$n $\backslash$n Answer: \\
\midrule
Texas & Webpage description: $\langle$ webpage description $\rangle$ $\backslash$n Question: Which of the following categories does this webpage belong to: 1) department, 2) project, 3) faculty, 4) course, 5) student, 6) other? If multiple options apply, provide a comma-separated list ordered from most to least related, then for each choice you gave, provide your reasoning. $\backslash$n $\backslash$n Answer: \\
\midrule
Wisconsin & Webpage description: $\langle$ webpage description $\rangle$ $\backslash$n Question: Which of the following categories does this webpage belong to: 1) department, 2) project, 3) faculty, 4) course, 5) student, 6) other? If multiple options apply, provide a comma-separated list ordered from most to least related, then for each choice you gave, provide your reasoning. $\backslash$n $\backslash$n Answer: \\
\midrule
Actor & Actor biography: $\langle$ actor description $\rangle$ $\backslash$n Question: Which of the following categories does this actor belong to: American film actors only, American film actors and American television actors, American television actors and American stage actors, English actors, Canadian actors? If multiple options apply, provide a comma-separated list ordered from most to least likely and provide your reasoning. $\backslash$n $\backslash$n Answer: \\
\midrule
Amazon-ratings & Product description: $\langle$ product description $\rangle$ $\backslash$n 
Question: Which of the following ratings does this product belong to: 1) Rating 5.0, 2) Rating 4.5, 
3) Rating 4.0, 4) Rating 3.5, 5) Rating lower than 3.5? If multiple options apply, provide a 
comma-separated list ordered from most to least related, then for each choice you gave, explain 
how it is present in the text. $\backslash$n $\backslash$n Answer:\\
\bottomrule
\end{tabular}
\caption{Prompts}
\label{tab:prompts}
\end{table*}

\section{Appendix C: Implementation Details}\label{app:implementation_details}

\subsection{Model Achitecture}
The whole training paradigm of is presented in Algorithm~\ref{alg:algorithm}.
\begin{algorithm}
\caption{The overall algorithm.}
\begin{algorithmic}[1]
\REQUIRE{Graph data $\mathcal{G=(V,E,T,Y)}$, model parameters $\theta$, maximum iterations $E$}
\ENSURE{Optimized model parameters $\theta^*$}
\STATE Enhance texts $\mathcal{T}' \gets \text{LLM}(\mathcal{T})$;
\FOR {$epoch = 0,1,...,E$}
\STATE Obtain node-level structure embeddings $ \mathbf{H}^{\zeta} \gets \text{GCN}(\mathcal{V,E})$, and neighborhood-level structure embeddings $\mathbf{H}^{\mathcal{N}}_i \gets \operatorname{Lin}( \operatorname{Concat}( \{ \mathbf{h}_k^{\zeta} \mid v_k \in \mathcal{N}_(v_i) \}))$;
\STATE Obtain token-level and sentence-level text embeddings $\mathbf{H}^{\varpi}, \mathbf{H}^t \gets \text{PLM}(\mathcal{T}')$;
\STATE Obtain the hierarchical similarity $\bar{\mathbf{S}}$ via Equation (1) and Equation (2);
\STATE Compute OT assignment $\mathbf{Q}^*$ from $\bar{\mathbf{S}}$ via Equation (3)--(5);
\STATE Perform alignment via Equation (7) -- (8);
\STATE Mining latent homophily via Equation (9) -- (10);
\STATE Compute total loss $\mathcal{L}$ via Equation (11), and update parameters $\theta$.
\ENDFOR
\RETURN{$\theta^*$}
\end{algorithmic}
\label{alg:algorithm}
\end{algorithm}

\subsection{Metric Definitions}\label{app:metrics}
We formally illustrate all metrics used in our empirical analysis. 
Following~\cite{Pei2020GeomGCN}, the node-level homophily $\mathcal{H}_N$, which measures the average proportion of neighbors sharing the same label, is calculated as
\begin{equation}
\begin{aligned}
H_N = \frac{1}{|\mathcal{V}|} \sum_{v_i \in \mathcal{V}} \frac{|\{v_j \in \mathcal{N}(i) \mid y_j = y_i \}|}{|\mathcal{N}(i)|}.
\end{aligned}
\end{equation}

Following~\cite{zhu2020H2GCN}, the edge-level homophily $\mathcal{H}_E$, which denotes the proportion of edges connecting nodes with the same label, is calculated as
\begin{equation}
\begin{aligned}
H_E = \frac{|\{(v_i,v_j) \in \mathcal{E} \mid y_i = y_j \}|}{|\mathcal{E}|}.
\end{aligned}
\end{equation}

The proportion of nodes whose all neighbors have same labels $\mathcal{R}_{\text{NYS}}$ is calculated as
\begin{equation}
\begin{aligned}
R_{\text{NYS}} = \frac{|\{v_i \in \mathcal{V} \mid \forall v_j \in \mathcal{N}(i),\ y_j = y_i \}|}{|\mathcal{V}|}.
\end{aligned}
\end{equation}

The proportion of nodes whose all neighbors have different labels $\mathcal{R}_{\text{NYD}}$ is calculated as
\begin{equation}
\begin{aligned}
R_{\text{NYD}} = \frac{|\{v_i \in \mathcal{V} \mid \forall v_j \in \mathcal{N}(i),\ y_j \neq y_i \}|}{|\mathcal{V}|}.
\end{aligned}
\end{equation}

Inspired by~\cite{luan2023CSBMPBE}, the proportion of nodes whose neighbors have mixed labels $\mathcal{R}_{\text{NYM}}$ is calculated as
\begin{equation}
\begin{aligned}
R_{\text{NYM}} = \frac{1}{|\mathcal{V}|} \sum_{v_i \in \mathcal{V}} \mathbb{I} (\exists v_j, 
v_k \in \mathcal{N}(i) : (y_j = y_i) \land (y_k \neq y_i)),
\end{aligned}
\end{equation}
where $\mathbb{I}(\cdot)$ is the indicator function. 

The proportion of nodes having unconnected same-label nodes $\mathcal{R}_{\text{UNYS}}$ is calculated as
\begin{equation}
\begin{aligned}
R_{\text{UNYS}} = \frac{|\{v_i \in \mathcal{V} \mid \exists v_j \in \mathcal{V} \setminus \mathcal{N}(i),\ y_j = y_i \}|}{|\mathcal{V}|}.
\end{aligned}
\end{equation}

The proportion of same-label pairs among unconnected pairs $\mathcal{R}_{\text{UEYS}}$ is calculated as
\begin{equation}
\begin{aligned}
R_{\text{UEYS}} = \frac{|\{(v_i,v_j) \notin \mathcal{E} \mid y_i = y_j \}|}{|\{(v_i,v_j) \notin 
\mathcal{E} \}|}.
\end{aligned}
\end{equation}

To measure the token-level semantic difference between neighboring nodes, we compute the complement of the average token-level similarity. 
\begin{equation}
\begin{aligned}
R_{\text{NWD}} = 1 - \mathbb{E}_{(v_i,v_j) \in \mathcal{E}} \left[ \operatorname{sim}(\mathbf{H}^{\mathcal{N}}_i, \mathbf{H}^{\mathcal{N}}_j) \right],
\end{aligned}
\end{equation}
where $\mathbb{E}$ is the average over edges$\mathcal{E}$. $\text{Sim}(\cdot,\cdot)$ denotes the cosine similarity, $\mathbf{H}^{\mathcal{N}}_i$ is token embedding matrix of node $v_i$. 

The semantic difference based on sentence-level embeddings is defined as
\begin{equation}
\begin{aligned}
R_{\text{NTD}} = 1 - \mathbb{E}_{(v_i,v_j) \in \mathcal{E}} \left[ \operatorname{sim}(\mathbf{h}^{t}_{i}, \mathbf{h}^{t}_{j}) \right],
\end{aligned}
\end{equation}
where $\operatorname{Sim}(\cdot,\cdot)$ denotes the cosine similarity function, and $\mathbf{h}^{t}_{i}$ denotes the text embedding of node $v_i$. 

To evaluate the semantic similarity among unconnected nodes, we measure the proportion of unconnected node pairs whose text embeddings exhibit high similarity, given a threshold $\tau$,
\begin{equation}
\begin{aligned}
R_{\text{UTS}} = \frac{|\{(v_i,v_j) \notin \mathcal{E} \mid \text{CosSim}(\mathbf{h}^{t}_{i}, \mathbf{h}^{t}_{j}) > \tau \}|}{|\{(v_i,v_j) \notin \mathcal{E} \}|},
\end{aligned}
\end{equation}
where $\tau$ is typically set to 0.5 following~\cite{wang2024understanding}.

\subsection{Prompts for Text Enhancement with LLM}\label{app:prompts}
Table~\ref{tab:prompts} outlines the prompts used for various datasets. Following TAPE~\citep{he2024TAPE}, each prompt includes the content and title of the text, followed by a task-specific question. The question is formulated to query the model about a particular aspect of the text and request an explanation for the prediction. The answer section is left blank for the model to fill in. Generally, our analysis finds that the current instructions allow the LLM to produce output that conforms well to the expected format without significant deviations, allowing the answers to be straightforwardly extracted from the text output of the LLM.

\subsection{Detailed Hyperparameters}

Table~\ref{tab:hyperparameters} provides an overview of the hyperparameters used for the GCN, GAT, SAGE, and DistillBERT models. These hyperparameters were selected based on the official OGB repository, and the language model hyperparameters follow those used in the repository. It is important to note that these hyperparameters were not tuned on a per-dataset basis, but instead were used consistently across all three TAG datasets based on those from prior work, and also set consistently across both our proposed method and the baselines. This demonstrates the generality and ease of use of our method, as well as its compatibility with existing GNN baselines.

\begin{table}
\small
\setlength{\tabcolsep}{1mm}
\centering
\begin{tabular}{l|cccc}
\toprule
Hyperparameters & GCN & GAT & SAGE & DistillBERT\\
\midrule
layers & 2 & 2 & 2 & 6 \\	
freeze layer	 & 0 & 0 & 0 & 3 \\	
hidden dimension & 768& 768 & 768 & 768\\
learning rate & 1e-2 & 1e-2 & 1e-2& 2e-5 \\
dropout & 0.5 & 0.5 & 0.5& 0.0\\
weight decay & 5e-4 & 5e-4 & 5e-4& 0.0\\
epoch	 	& 1000 & 1000 & 1000& 1000 \\
early stop & 50 & 50 & 50 & 50 \\
batch size & NA & NA & NA & 512 \\
warmup ratio& NA & NA & NA & 0.1\\
\bottomrule
\end{tabular}
\caption{Hyperparameters}
\label{tab:hyperparameters}
\end{table}

In practice, the prompt value $p$ is adaptively set as the bottom 10\% percentile of all original similarity values, providing a data-driven and flexible filtering threshold. We adopt mini-batch training based on subgraph sampling, with a batch size of 512. These results may be lower than normal because subgraph sampling will lose global information.

\section{Appendix D: Experimental data and results}\label{app:experiments}

\begin{table*}[htbp]
\centering
\small
\setlength{\tabcolsep}{0.9mm}
\begin{tabular}{lccccccccccccc}
\toprule
Dataset & Node & Edge & Feature & Class
& $H_N$ & $H_E$ & $R_{NYS}$ & $R_{NYD}$ & $R_{NYM}$
& $R_{UNS}$ & $R_{NWD}$ & $R_{NTD}$ & $R_{UTS}$ \\
\midrule
Cora & 2,708 & 5,278 & 1,433 & 7
& 82.52 & 81.00 & 65.58 & 5.80 & 28.62 & 99.17 & 85.77 & 18.22 & 82.98 \\
Pubmed & 19,717 & 44,324 & 500 & 3
& 79.24 & 80.20 & 64.43 & 12.79 & 22.78 & 99.95 & 81.12 & 13.71 & 94.80 \\
Products & 54,025 & 74,420 & 100 & 47
& 78.97 & 80.80 & 70.11 & 16.21 & 13.68 & 99.95 & 94.89 & 29.72 & 54.19 \\
arXiv & 169,343 & 1,166,243 & 128 & 40
& 63.53 & 65.50 & 28.35 & 12.28 & 59.36 & 99.93 & 80.09 & 23.59 & 19.25 \\
Amazon-ratings & 24,492 & 186,100 & 300 & 5
& 37.57 & 38.00 & 2.55 & 16.08 & 81.37 & 99.95 & 76.70 & 17.06 & 85.53 \\
arXiv23 & 46,198 & 78,548 & 128 & 40
& 29.66 & 64.70 & 54.15 & 19.31 & 26.53 & 99.98 & 83.87 & 17.93 & 51.34 \\
Wisconsin & 265 & 530 & 1,703 & 5
& 18.68 & 21.90 & 4.53 & 60.38 & 35.09 & 98.83 & 88.06 & 16.59 & 95.67 \\
Cornell & 195 & 304 & 1,703 & 5
& 12.12 & 13.20 & 4.62 & 74.36 & 21.03 & 99.06 & 89.54 & 18.87 & 96.08 \\
Texas & 187 & 328 & 1,703 & 5
& 10.68 & 10.80 & 3.74 & 77.54 & 18.72 & 99.56 & 89.52 & 18.88 & 97.63 \\
Actor & 4,416 & 12,172 & 932 & 5
& 60.86 & 56.00 & 34.40 & 16.19 & 49.41 & 99.76 & 72.74 & 23.75 & 77.57 \\
\bottomrule
\end{tabular}
\caption{Dataset statistics and performance of all methods.}
\label{tab:merged_all}
\end{table*}
\subsection{Dataset and Heterophilic Metrics Results}\label{app:dataset}

\begin{itemize}
\item Cora, Pubmed~\cite{yang2016revisiting, sen2008collective_pubmed}, arXiv~\cite{hu2020ogb}, and arXiv23~\cite{he2024TAPE} are four citation networks that are considered classic homophilic graphs. In these graphs, nodes represent papers, edges represent citation relationships between two papers, and labels indicate the research topic of each paper. 
\item Products~\cite{hu2020ogb} is a co-purcharse network. In this graph, nodes represent products, edges represent co-purchase relationships between two products, and labels indicate the product category of each product. 
\item Cornell, Texas, and Wisconsin~\cite{Pei2020GeomGCN} are derived from computer science department websites where webpages serve as nodes, hyperlinks as edges, and entity types as labels. 
\item Actor~\cite{tang2009social} is an actor co-occurrence network with nodes representing actors, edges denoting their collaborations, and labels indicating the Wikipedia-derived actor categories of each actor.
\item Amazon-ratings~\cite{platonov2023critical} is a product co-purchasing network where nodes represent products, edges link products frequently bought together, and labels correspond to rating levels assigned to each product.
\end{itemize}

\subsection{Additional Results on Node Classification.}
\begin{table}[htbp]
\centering
\begin{tabular}{lc}
\toprule
Method & Actor \\
\midrule
MLP & 71.44$\pm$1.3 \\
Distilbert & 75.44$\pm$0.7 \\
GPT3.5 & 67.41 \\
GCN & 73.98$\pm$0.7 \\
GAT & 66.43$\pm$1.3 \\
SAGE & 74.11$\pm$1.1 \\
H2GCN & 74.97$\pm$1.2 \\
FAGCN & 73.39$\pm$1.0 \\
TAPE (GCN) & 81.36$\pm$0.4 \\
\midrule
Ours-GCN & 81.56$\pm$1.3\\
\bottomrule
\end{tabular}
\caption{Node classification accuracy on Actor.}
\label{tab:actor}
\end{table}

\subsection{OT Assignment Visualization}
We further examine the OT assignment matrix in three scenarios (partial heterophily, complete heterophily, and latent homophily). Visualizations include comparisons between OT allocations, adjacency matrices, semantic similarities, and label distributions to highlight our alignment strategy's interpretability and effectiveness.

\begin{figure}[htbp]
\centering
\begin{subfigure}[b]{0.4\linewidth}
\centering
\includegraphics[width=\linewidth]{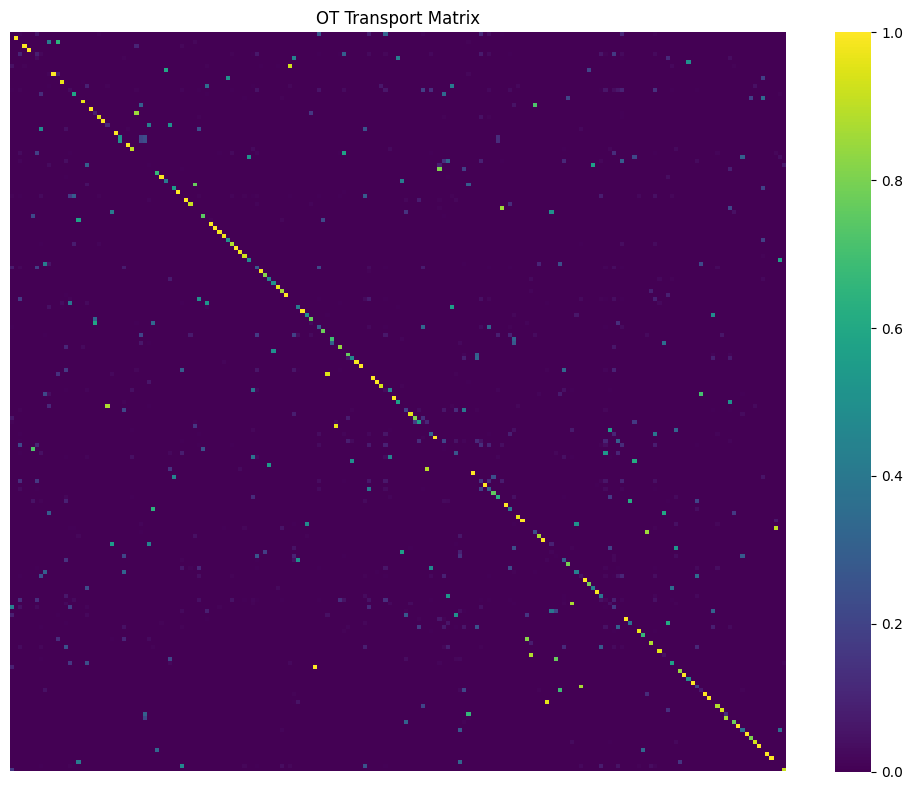}
\caption{OT Assignment}
\label{fig:OT}
\end{subfigure}
\begin{subfigure}[b]{0.4\linewidth}
\centering
\includegraphics[width=\linewidth]{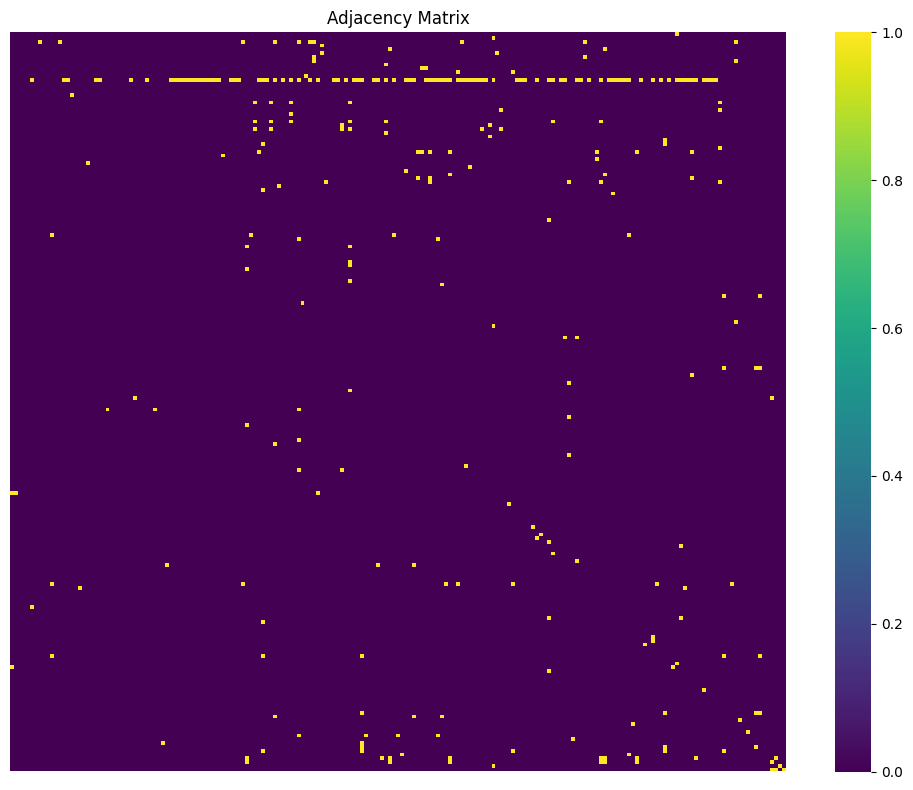}
\caption{Adjacency Matrix}
\label{fig:Adjacency_matrix}
\end{subfigure}
\\
\begin{subfigure}[b]{0.4\linewidth}
\centering
\includegraphics[width=\linewidth]{images/cornell_adj.png}
\caption{Senmatic similarity}
\label{fig:Senmatic_similarity_matrix}
\end{subfigure}
\begin{subfigure}[b]{0.4\linewidth}
\centering
\includegraphics[width=\linewidth]{images/cornell_adj.png}
\caption{label distribution}
\label{fig:label_distribution_matrix}
\end{subfigure}
\caption{OT Assignment Visualization.}
\label{fig:ot_assignment_visualization}
\end{figure}

\begin{figure*}[hbp]
\centering
\begin{subfigure}[b]{0.19\linewidth}
\centering
\includegraphics[width=1.05in,height=0.867in]{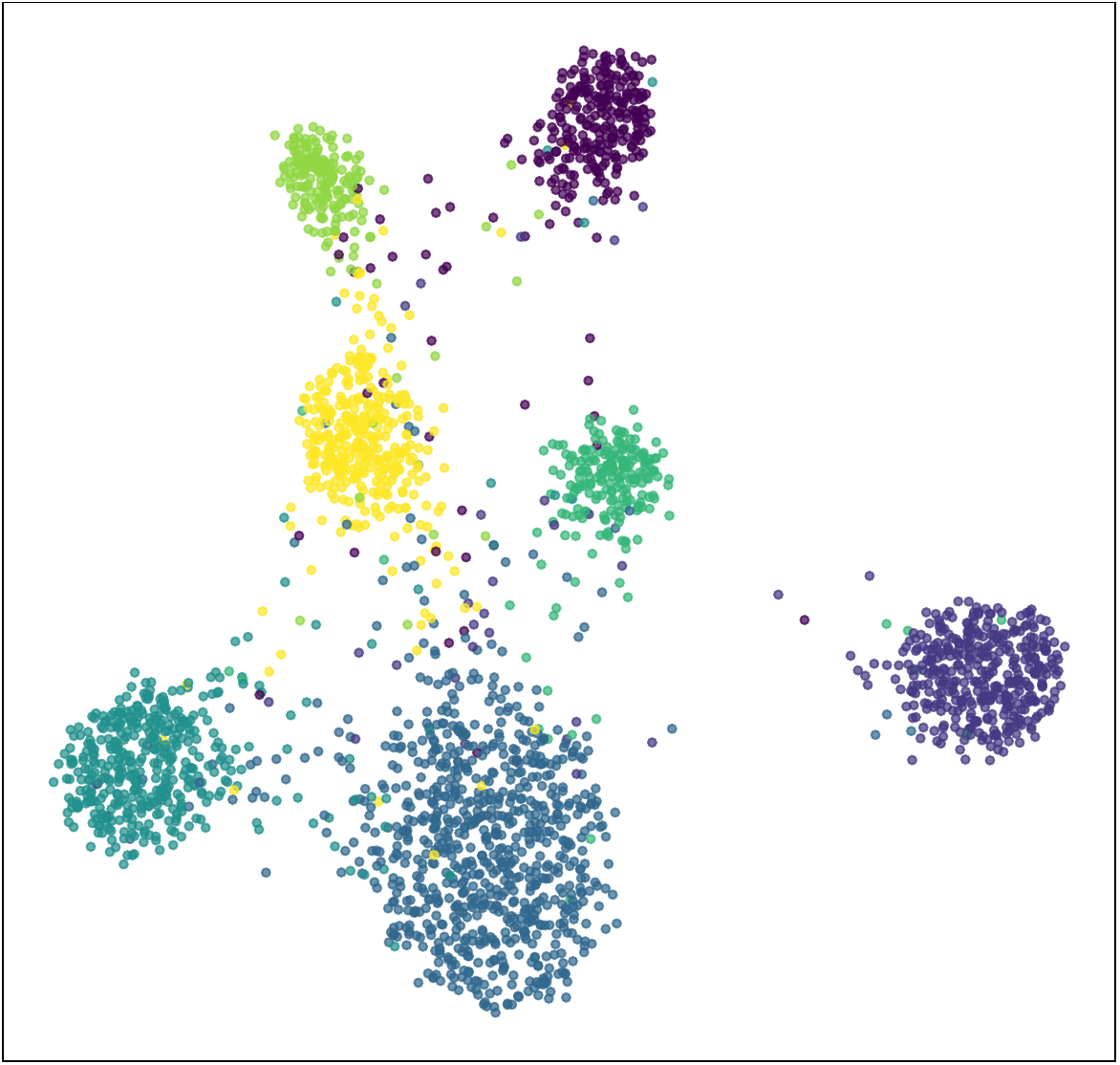}
\caption{Cora}
\label{fig:Cora}
\end{subfigure}
\begin{subfigure}[b]{0.19\linewidth}
\centering
\includegraphics[width=1.05in,height=0.867in]{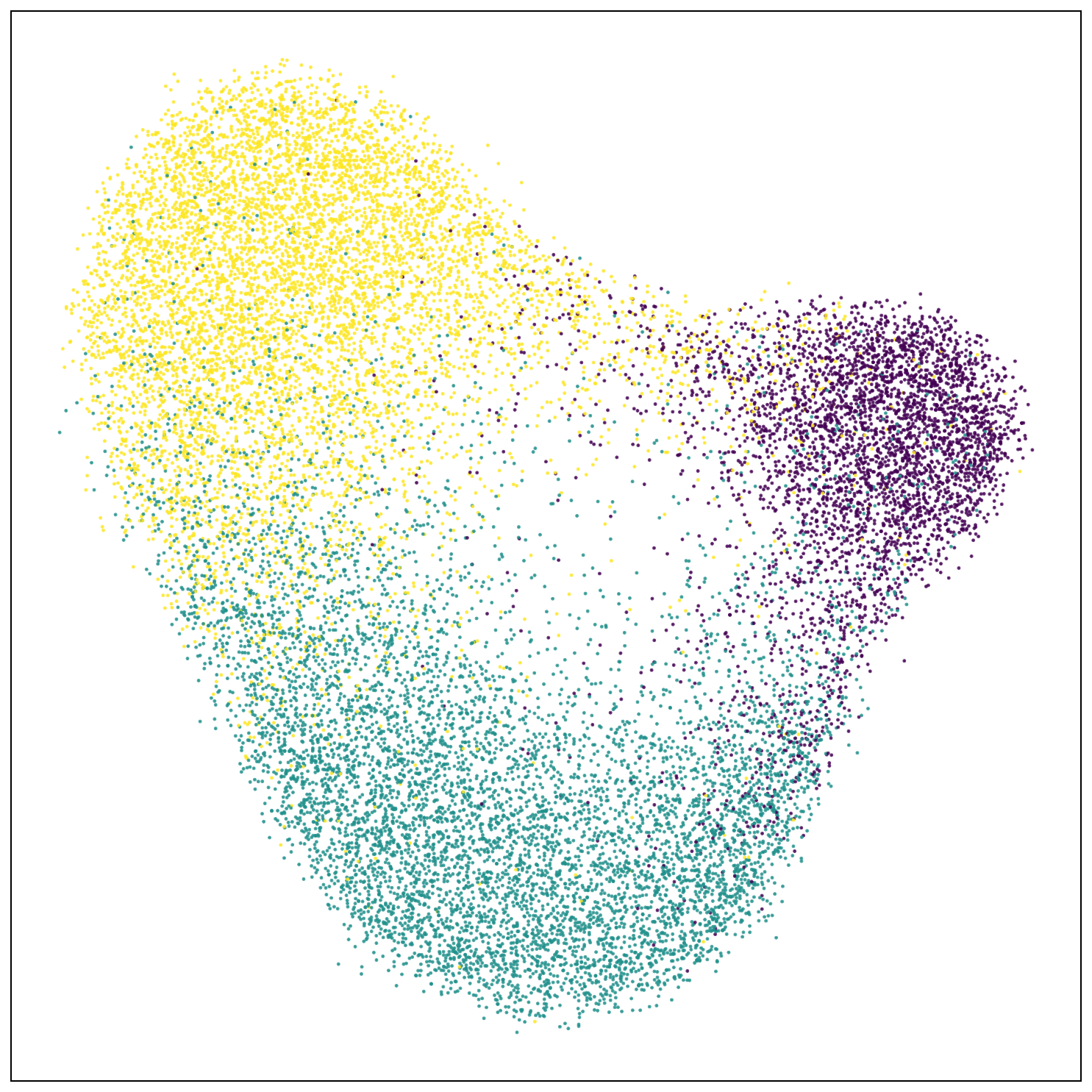}
\caption{PubMed}
\label{fig:PubMed}
\end{subfigure}
\begin{subfigure}[b]{0.19\linewidth}
\centering
\includegraphics[width=1.05in,height=0.867in]{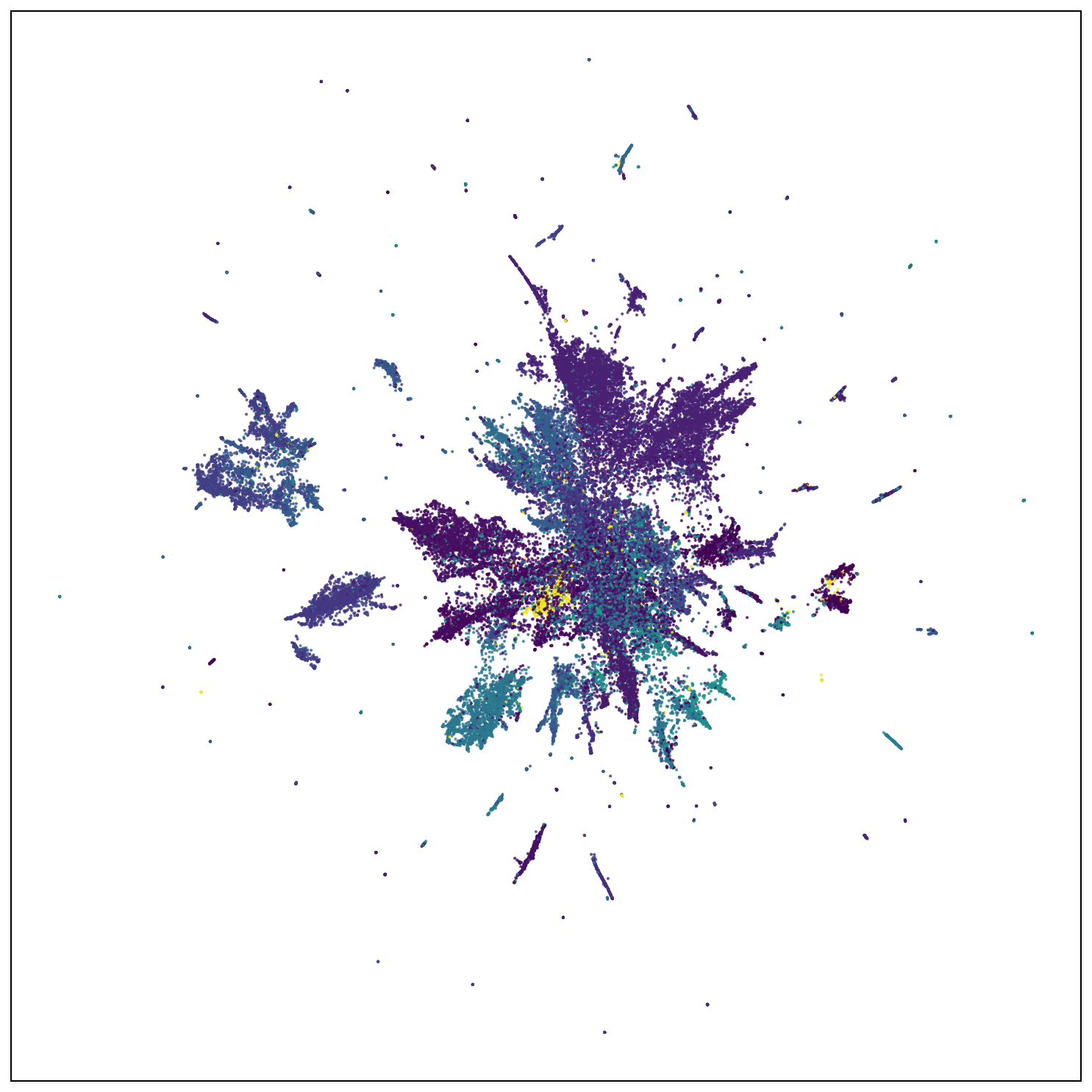}
\caption{Products}
\label{fig:Products}
\end{subfigure}
\begin{subfigure}[b]{0.19\linewidth}
\centering
\includegraphics[width=1.05in,height=0.867in]{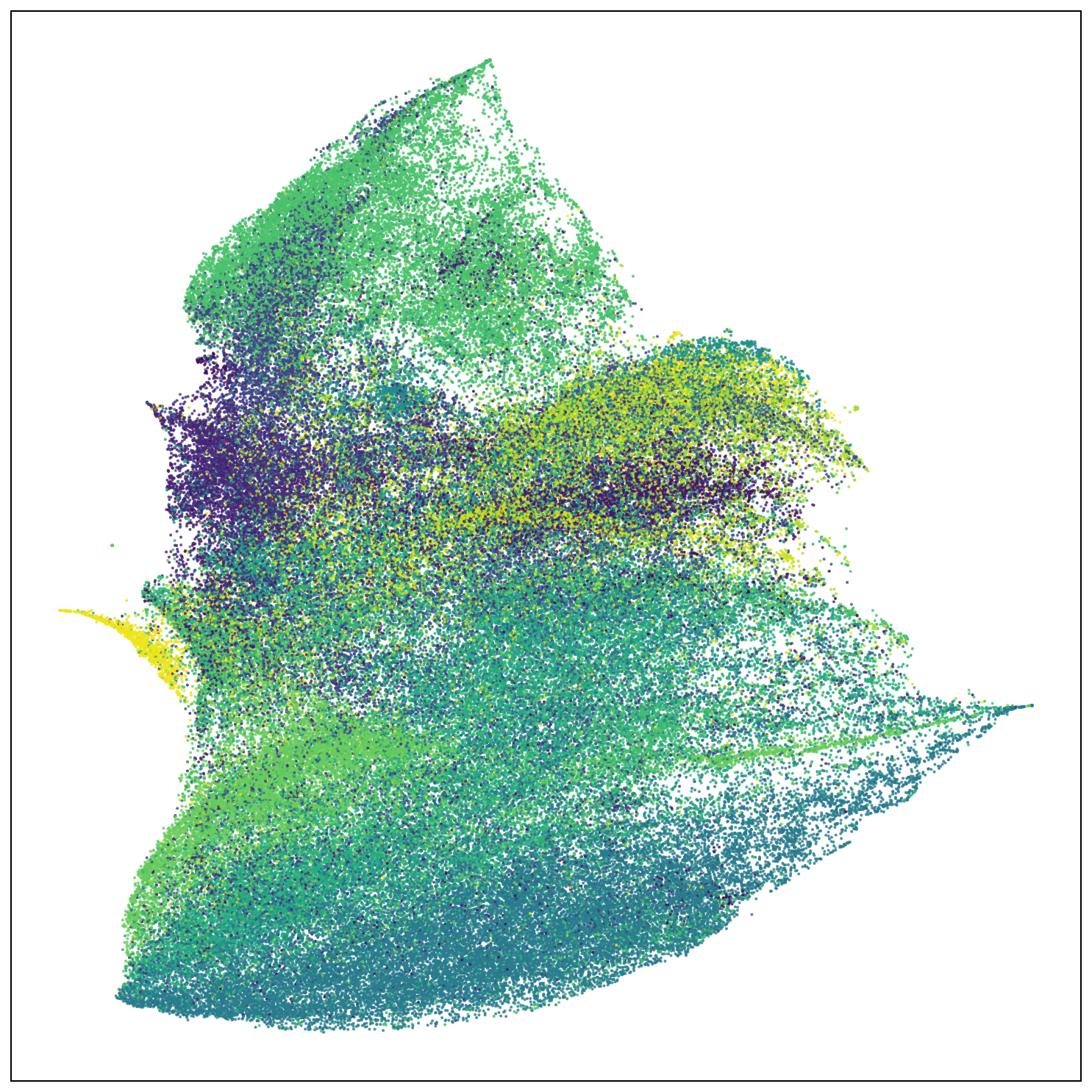}
\caption{arXiv}
\label{fig:arXiv}
\end{subfigure}
\begin{subfigure}[b]{0.19\linewidth}
\centering
\includegraphics[width=1.05in,height=0.867in]{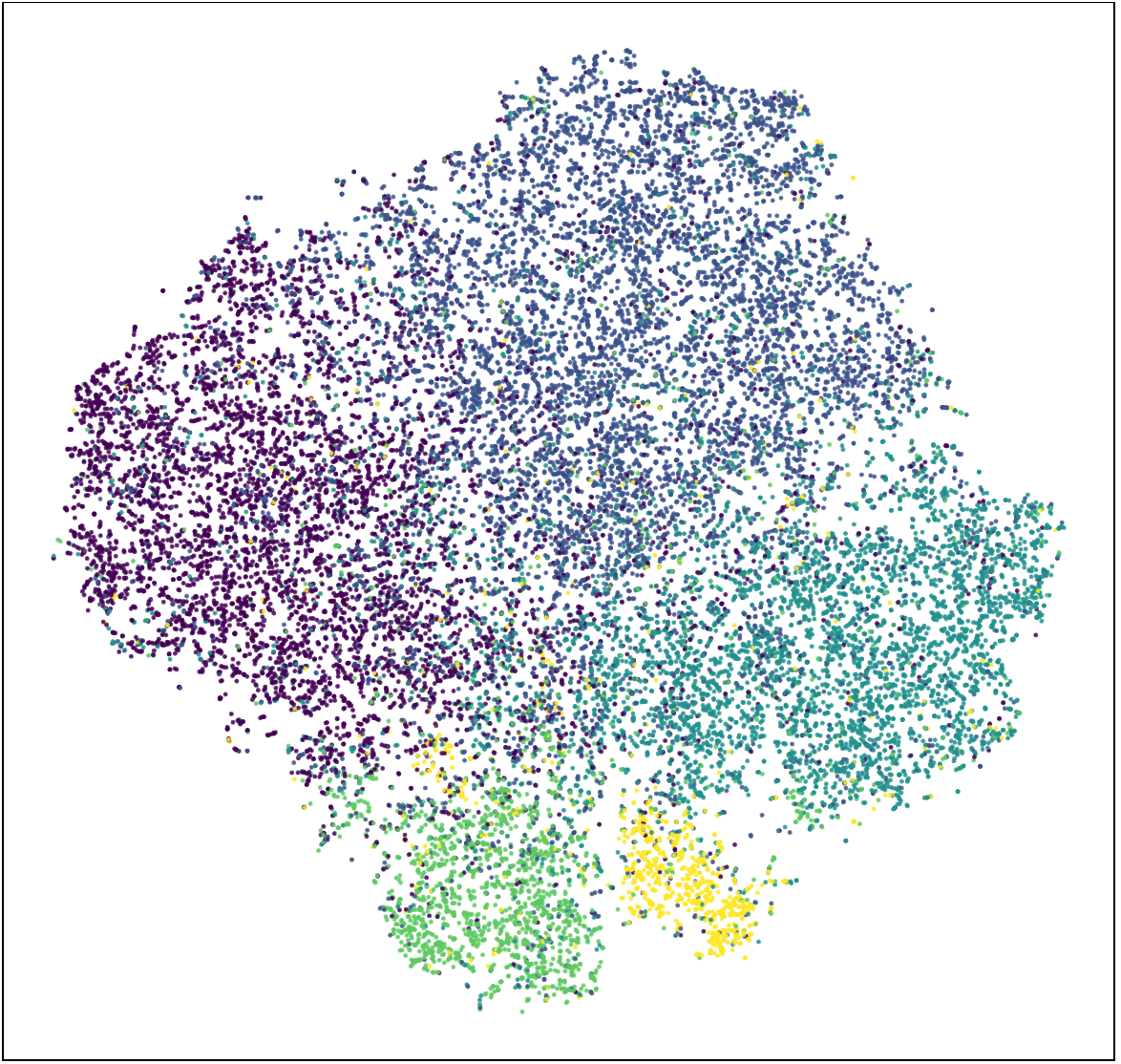}
\caption{Amazon}
\label{fig:Amazon}
\end{subfigure}
\\
\begin{subfigure}[b]{0.19\linewidth}
\centering
\includegraphics[width=1.05in,height=0.867in]{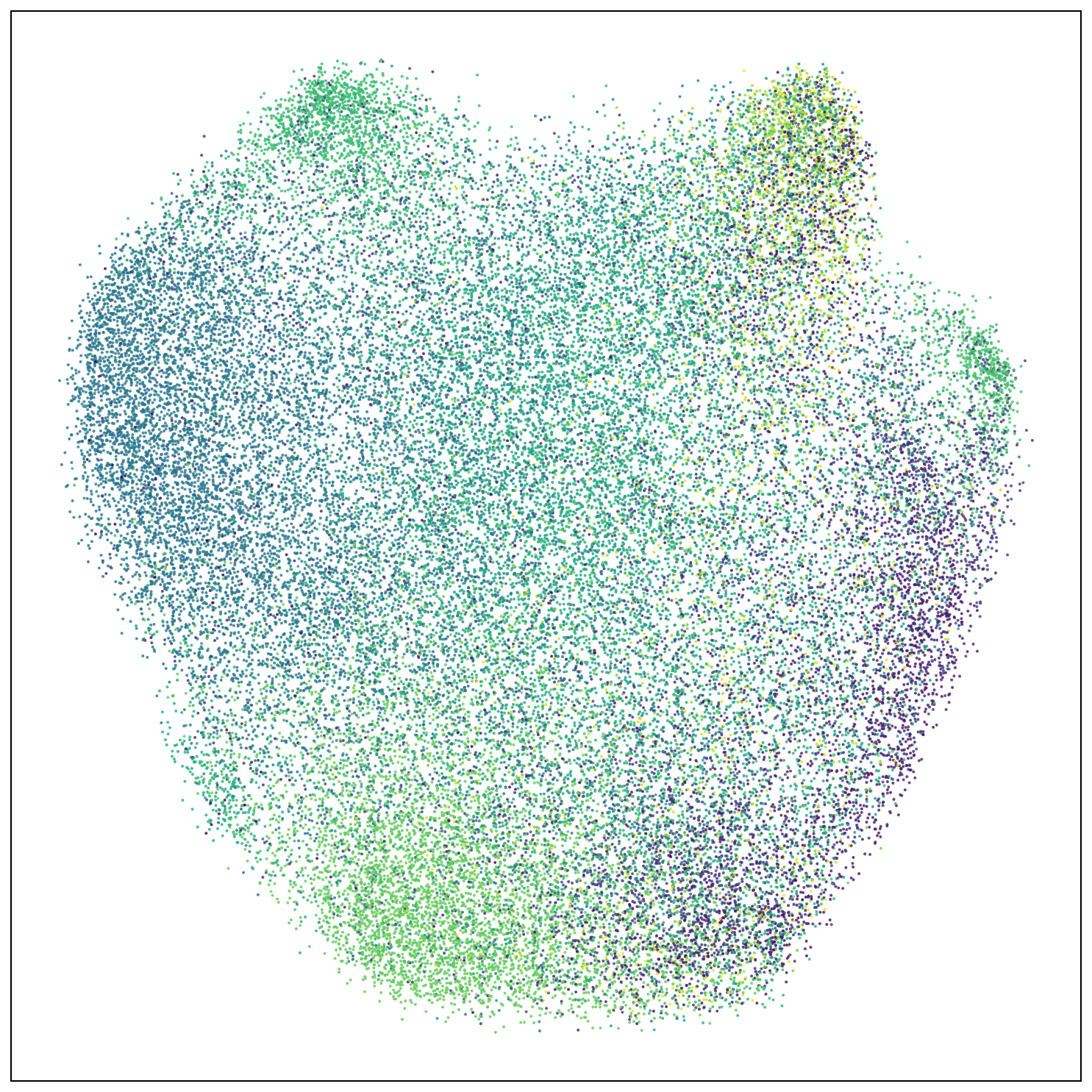}
\caption{arXiv23}
\label{fig:arXiv23}
\end{subfigure}
\begin{subfigure}[b]{0.19\linewidth}
\centering
\includegraphics[width=1.05in,height=0.867in]{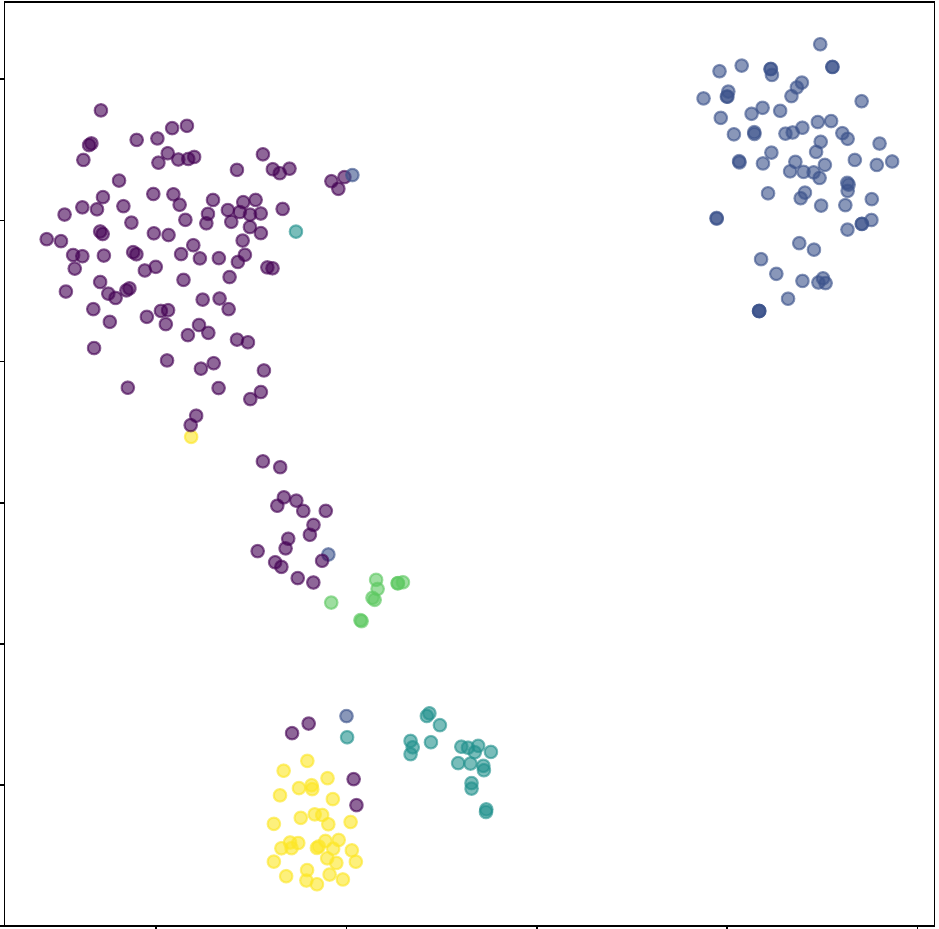}
\caption{Wisconsin}
\label{fig:Wisconsin}
\end{subfigure}
\begin{subfigure}[b]{0.19\linewidth}
\centering
\includegraphics[width=1.05in,height=0.867in]{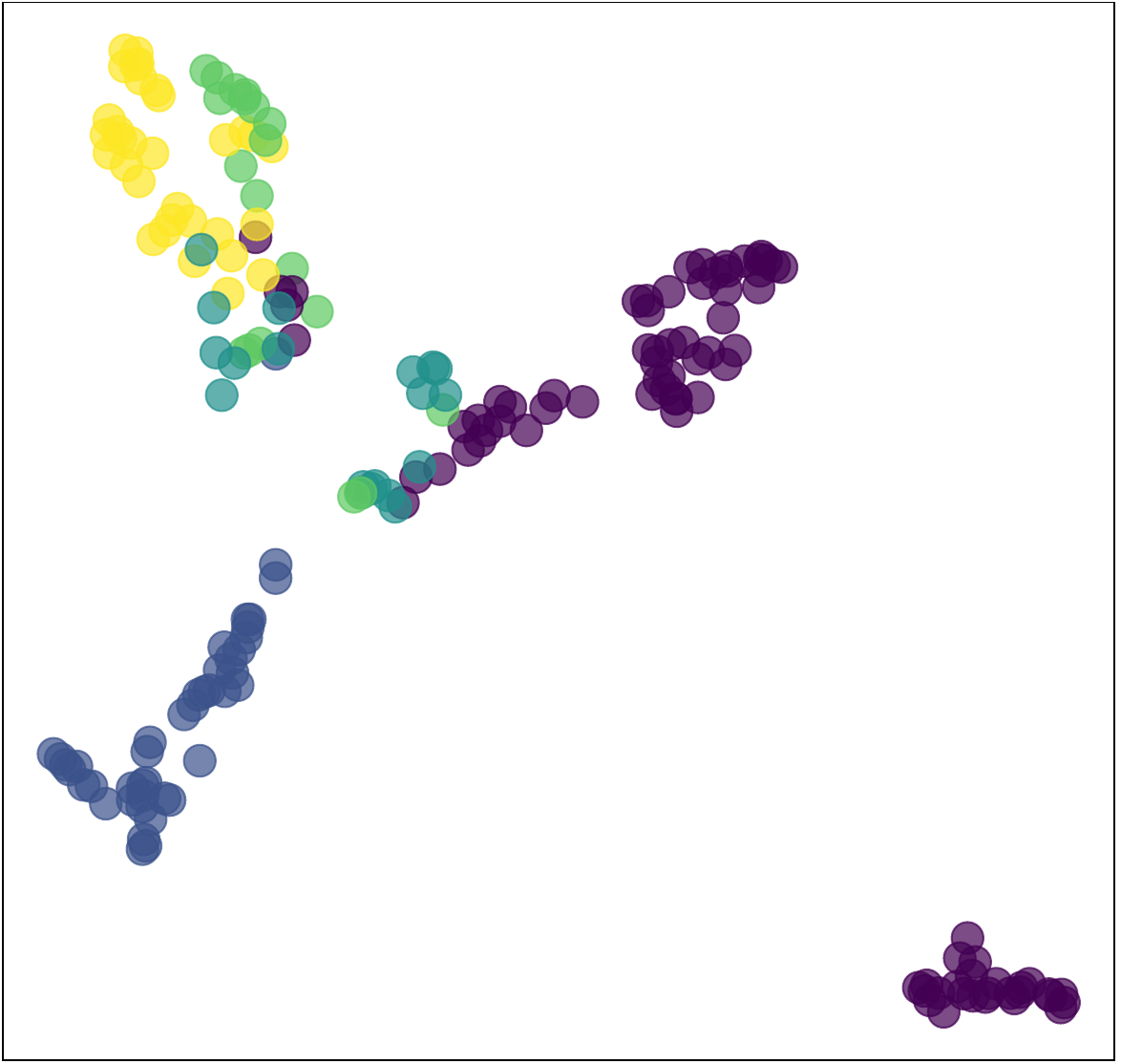}
\caption{Cornell}
\label{fig:Cornell}
\end{subfigure}
\begin{subfigure}[b]{0.19\linewidth}
\centering
\includegraphics[width=1.05in,height=0.867in]{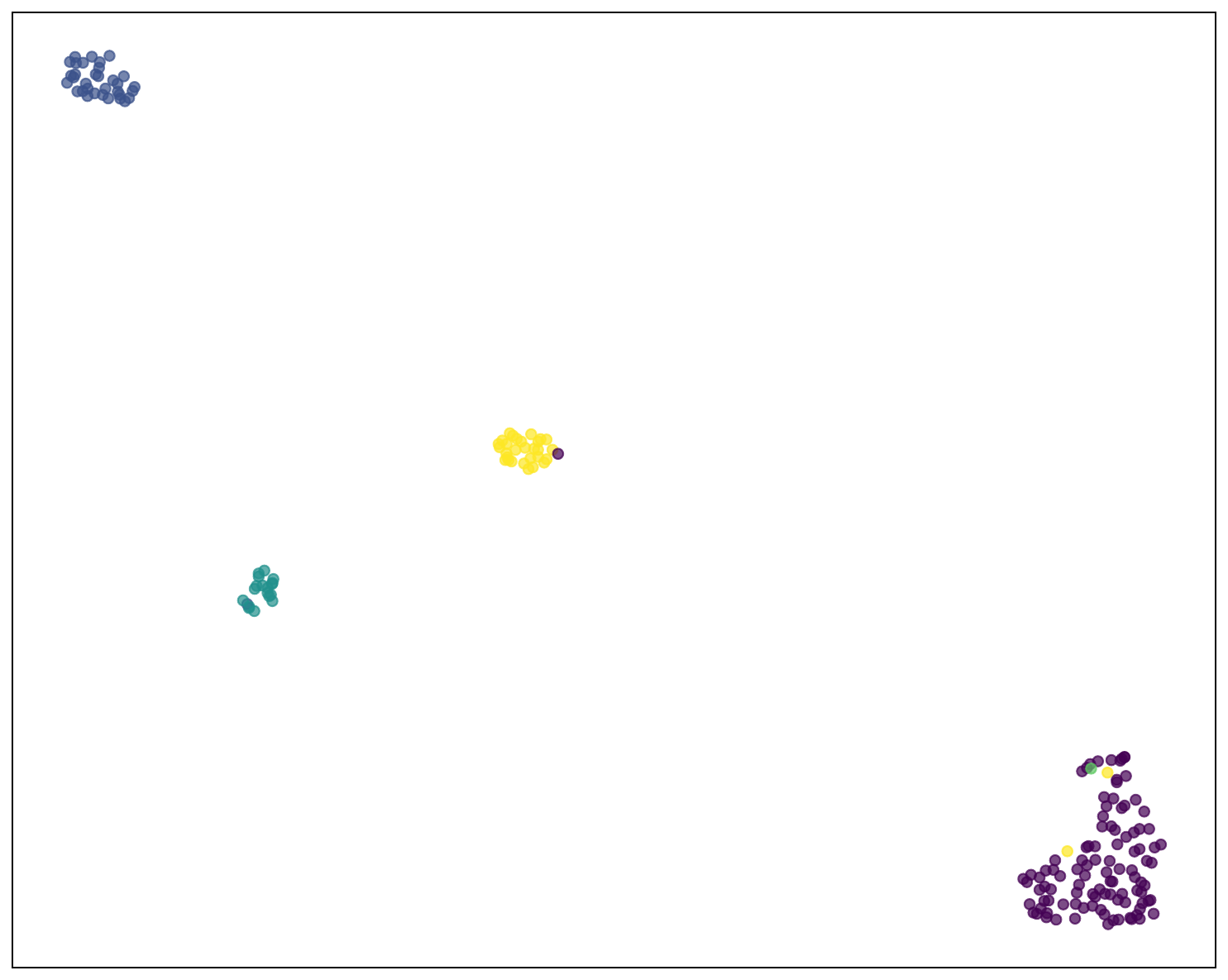}
\caption{Texas}
\label{fig:Texas}
\end{subfigure}
\begin{subfigure}[b]{0.19\linewidth}
\centering
\includegraphics[width=1.05in,height=0.867in]{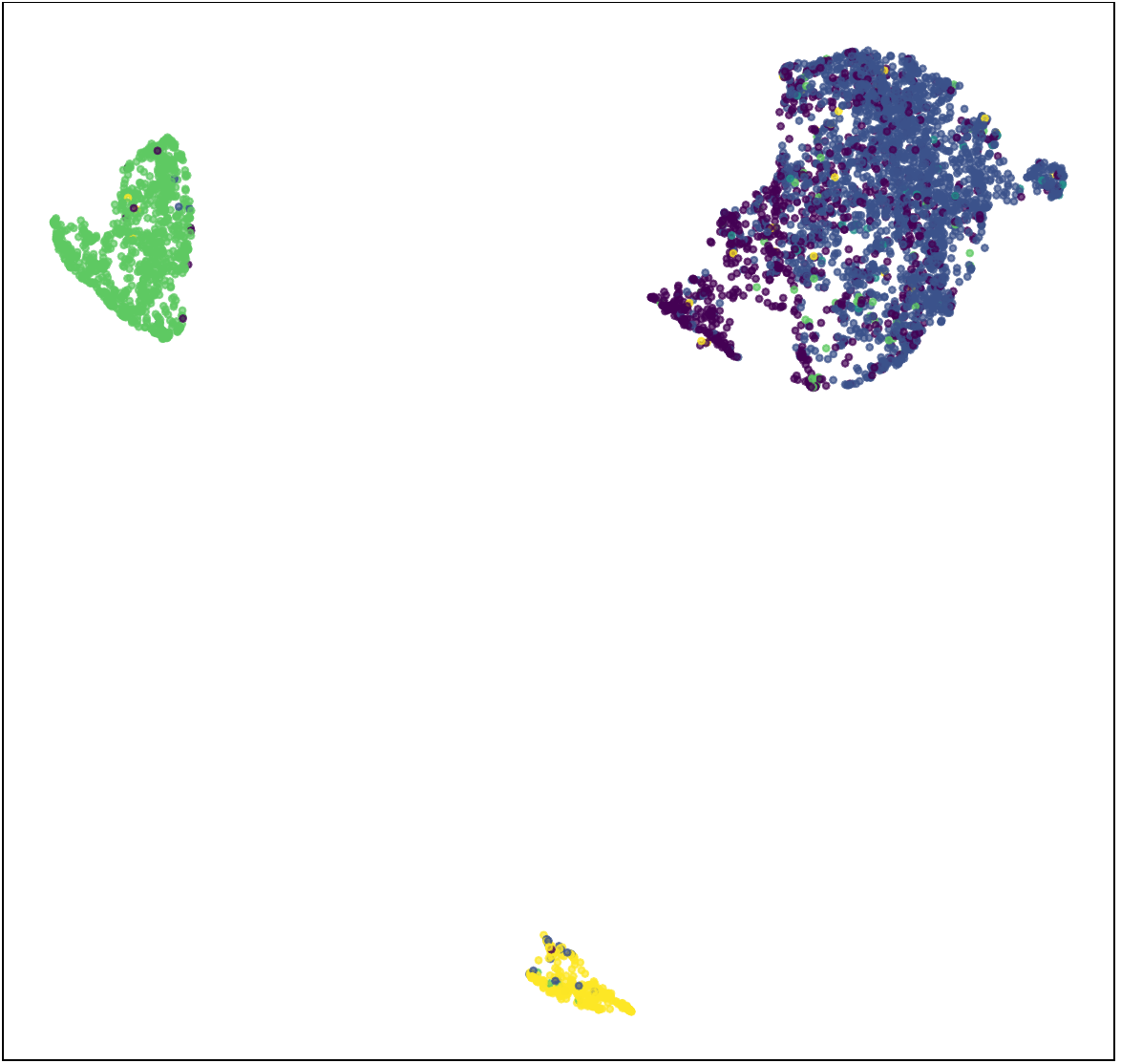}
\caption{Actor}
\label{fig:Actor}
\end{subfigure}
\caption{T-SNE visualization on different datasets (seed = 0), with colors indicating ground-truth class labels.}
\label{fig:tsne1}
\end{figure*}

\subsection{Additional Visualization of Learned Representations.}
We provide t-SNE visualizations of the learned node embeddings to qualitatively assess the embedding space. Results across the datasets are shown in Figure~\ref{fig:tsne1} (seed = 0), where each point represents a node and colors denote ground-truth classes.

Across datasets, several plots show compact, well-separated clusters, while others display overlapping or elongated manifolds—consistent with varying degrees of label homophily and heterophily.

\end{document}